\documentclass{article}
\usepackage{iclr2027_conference,times}

\usepackage{amsmath,amsfonts,bm}

\def\eqref#1{equation~\ref{#1}}

\def\1{\bm{1}}

\DeclareMathAlphabet{\mathsfit}{\encodingdefault}{\sfdefault}{m}{sl}
\SetMathAlphabet{\mathsfit}{bold}{\encodingdefault}{\sfdefault}{bx}{n}

\usepackage{algorithm}
\usepackage{algpseudocode}
\usepackage{hyperref}
\usepackage{url}
\usepackage{booktabs}
\usepackage{graphicx}
\usepackage{colortbl}
\usepackage{multirow}
\usepackage{arydshln}
\usepackage{wrapfig}
\usepackage{float}
\usepackage{listings}
\usepackage{tcolorbox}
\tcbuselibrary{listings,breakable,skins}
\definecolor{artifactBackground}{HTML}{F8FAFC}
\definecolor{artifactBorder}{HTML}{2F4F6F}
\definecolor{artifactExcerpt}{HTML}{1E4F8A}
\lstdefinestyle{harnessSource}{%
  basicstyle=\fontsize{8.5}{10.5}\fontencoding{OT1}\fontfamily{cmtt}\fontseries{m}\fontshape{n}\selectfont,
  columns=fullflexible,
  keepspaces=true,
  breaklines=true,
  breakatwhitespace=true,
  breakindent=0pt,
  breakautoindent=false,
  postbreak={},
  showstringspaces=false,
  aboveskip=0pt, belowskip=0pt,
  tabsize=4,
}
\lstdefinestyle{harnessText}{%
  style=harnessSource,
  breakindent=0pt,
  postbreak={},
  escapeinside={(*@}{@*)},
}
\newtcblisting[auto counter]{artifactbox}[2][]{%
  breakable, enhanced,
  listing only,
  listing options={style=harnessSource},
  colback=artifactBackground,
  colframe=artifactBorder,
  boxrule=0.6pt,
  arc=4pt, outer arc=4pt,
  left=8pt, right=7pt, top=5pt, bottom=5pt,
  boxsep=1pt,
  before skip=1em, after skip=0.7em,
  colbacktitle=artifactBorder,
  coltitle=white,
  fonttitle=\fontencoding{OT1}\fontfamily{cmr}\bfseries\footnotesize,
  toptitle=3pt, bottomtitle=3pt,
  title={Listing~\thetcbcounter\quad #2},
  title after break={Listing~\thetcbcounter\quad #2 (continued)},
  label={#1},
}

\newtcblisting[use counter from=artifactbox]{artifacttextbox}[2][]{%
  breakable, enhanced,
  listing only,
  listing options={style=harnessText},
  colback=artifactBackground,
  colframe=artifactBorder,
  boxrule=0.6pt,
  arc=4pt, outer arc=4pt,
  left=8pt, right=7pt, top=5pt, bottom=5pt,
  boxsep=1pt,
  before skip=1em, after skip=0.7em,
  colbacktitle=artifactBorder,
  coltitle=white,
  fonttitle=\fontencoding{OT1}\fontfamily{cmr}\bfseries\footnotesize,
  toptitle=3pt, bottomtitle=3pt,
  title={Listing~\thetcbcounter\quad #2},
  title after break={Listing~\thetcbcounter\quad #2 (continued)},
  label={#1},
}

\newtcblisting[auto counter]{diagnosisbox}[2][]{%
  breakable, enhanced,
  listing only,
  listing options={style=harnessText},
  colback=artifactBackground,
  colframe=artifactBorder,
  boxrule=0.6pt,
  arc=4pt, outer arc=4pt,
  left=8pt, right=7pt, top=5pt, bottom=5pt,
  boxsep=1pt,
  before skip=1em, after skip=0.7em,
  colbacktitle=artifactBorder,
  coltitle=white,
  fonttitle=\fontencoding{OT1}\fontfamily{cmr}\bfseries\footnotesize,
  toptitle=3pt, bottomtitle=3pt,
  title={Case~\thetcbcounter\quad #2},
  title after break={Case~\thetcbcounter\quad #2 (continued)},
  label={#1},
}

\newcommand{\method}{\textsc{HarnessPro}}

\title{HarnessPro: Recursive Harness Self-Improvement with Pareto Optimization for Performance-Cost Trade-offs}

\author{
\textbf{\href{https://openreview.net/profile?id=~Zhuoshi_Pan2}{Zhuoshi Pan}}$^{1,*}$,
\textbf{\href{https://openreview.net/profile?id=~ZhiYuan_Feng1}{ZhiYuan Feng}}$^{1,*}$,
\textbf{\href{https://openreview.net/profile?id=~Yukuo_Cen1}{Yukuo Cen}}$^{2}$,
\textbf{\href{https://openreview.net/profile?id=~Honglin_Lin2}{Honglin Lin}}$^{3}$,\\
\textbf{\href{https://openreview.net/profile?id=~Junru_Lu1}{Junru Lu}}$^{4}$,
\textbf{\href{https://openreview.net/profile?id=~Zhongxia_Chen2}{Zhongxia Chen}}$^{5}$,
\textbf{\href{https://openreview.net/profile?id=~Shuo_Yang45}{Shuo Yang}}$^{5}$,
\textbf{\href{https://openreview.net/profile?id=~H._Vicky_Zhao1}{H. Vicky Zhao}}$^{1,\dagger}$,
\textbf{\href{https://openreview.net/profile?id=~Jie_Tang1}{Jie Tang}}$^{1,\dagger}$\\
\textsuperscript{1}\href{https://tsinghua.edu.cn}{Tsinghua University}\quad
\textsuperscript{2}\href{https://z.ai}{Z.AI}\quad
\textsuperscript{3}\href{https://www.sjtu.edu.cn}{Shanghai Jiao Tong University}\\
\textsuperscript{4}\href{https://warwick.ac.uk}{University of Warwick}\quad
\textsuperscript{5}\href{https://klingai.com}{Kling AI}\\
{\small\normalfont $^{*}$Equal contribution.\quad$^{\dagger}$Corresponding authors.}
}

\iclrfinalcopy
\begin{document}

\maketitle
\lhead{Under review}

\begin{abstract}
Harness development plays a crucial role in building capable AI agents.
Recent approaches improve harnesses through recursive self-improvement (RSI), but their performance gains often come with higher execution costs.
We introduce \method{}, a label-free method for evolving harnesses to improve task performance while reducing execution cost.
In this setting, candidate harnesses are evaluated using LLM-generated tests without access to ground-truth solutions or official tests.
Specialized agents diagnose execution trajectories, plan updates with different performance-cost priorities, and implement them as candidate harnesses.
Rather than returning a single harness, \method{} uses normalized weighted squared-distance selection to provide \textbf{cost-focused} (lite), \textbf{balanced} (medium), and \textbf{performance-focused} (heavy) harnesses for different deployment budgets.
Selected harnesses enter the next iteration, where execution feedback guides further updates.
Our experiments cover Terminal-Bench~2, SWE-Bench Pro, and ProgramBench, using GPT-5.5 with Codex and GLM-5.2 with Claude Code.
Across these benchmarks using GPT-5.5 with Codex, the heavy mode of \method{} achieves an average pass-rate gain of 2.0 percentage points and an average per-task cost reduction of 26.1\% relative to RHO.
This advantage also holds when using GLM-5.2 with Claude Code.
\mbox{Code is available at \href{https://github.com/pzs19/HarnessPro}{this link}.}
\end{abstract}

\section{Introduction}
\label{sec:introduction}

An agent harness is the software infrastructure that manages interactions between a large language model (LLM) and its environment~\citep{grace2026demystifying}.
Recent work on recursive self-improvement (RSI)~\citep{zhang2026dgm} uses past execution trajectories to diagnose problems and guide harness updates that improve performance on future tasks~\citep{suzgun2026dynamiccheatsheet,lin2026ahe,pan2026rho}.
However, these methods often pursue performance gains through more extensive exploration and stronger verification~\citep{lin2026ahe,pan2026rho}.
These efforts improve performance, but often at the cost of higher token consumption and increased latency.
This trade-off is less appealing in practical deployment, where the execution cost and latency also matter.
We therefore ask: \emph{\textbf{Can recursive harness self-improvement deliver better performance at lower execution cost?}}

Toward this goal, we introduce \method{}, a label-free method for recursive harness self-improvement that uses both performance and cost metrics to guide harness updates and candidate selection.
It orchestrates several specialized agents to turn execution feedback into successive harness updates.
A trajectory analyzer examines repeated rollouts to identify failures and avoidable overhead.
Guided by different \textbf{performance-cost priorities}, a planner turns these diagnoses into harness improvement plans, which an editor implements as executable harness candidates.
A test-synthesis agent provides fixed tests for evaluating the candidates without access to ground-truth solutions or official tests.
Candidates that underperform the initial harness or are Pareto-dominated are discarded.
To accommodate different budgets, we then use \textit{normalized weighted squared-distance selection}~\citep{marler2004survey} to retain representative harnesses for \textbf{cost-focused} (lite), \textbf{balanced} (medium), and \textbf{performance-focused} (heavy) modes.
The selected harnesses then enter the next epoch, where further updates pursue different performance-cost priorities.

We evaluate \method{} on Terminal-Bench~2~\citep{tbench2025}, SWE-Bench Pro~\citep{deng2026swebenchpro}, and ProgramBench~\citep{yang2026programbench}, using GPT-5.5 with Codex and GLM-5.2 with Claude Code.
With both backbone models, \method{} \textbf{improves task performance} while \textbf{reducing execution cost}.
For example, on Terminal-Bench~2 with GPT-5.5 and Codex, the heavy mode of \method{} outperforms RHO~\citep{pan2026rho} in pass rate by 4.5 percentage points while reducing mean cost per task by 33.9\%.
For more cost-sensitive deployments, the medium mode delivers performance comparable to RHO at substantially lower execution cost.

An analysis of the evolved harnesses shows that \method{} does not improve agents simply by compressing context or increasing reasoning effort.
The evolved harnesses \textbf{\textit{dynamically}} allocate \textbf{\textit{reasoning effort}} and \textbf{\textit{tool budgets}} based on task descriptions, while runtime hooks steer execution from exploration toward convergence based on the assigned budgets.
They also adopt tool-output compression like RTK~\citep{szymkowiak2026rtk} and streamline instructions to reduce redundant prompt content, while retaining necessary context and final verification.
Together, these mechanisms curb \textbf{\textit{avoidable overhead}} while preserving the reasoning and verification needed for task success, thereby advancing the performance-cost \textbf{\textit{Pareto}} frontier.
Our contributions are threefold:
\begin{itemize}
\item We introduce \method{}, a multi-agent framework that simultaneously optimizes performance and cost by analyzing execution trajectories and recursively improving harnesses.
\item We assess candidate artifacts with fixed LLM-generated tests rather than direct LLM judgments, improving failure detection without access to official tests.
\item Across three benchmarks, \method{} improves performance while reducing cost and latency, yielding a series of harnesses with different performance-cost trade-offs.
\end{itemize}

\section{Related Work}
\label{sec:related-work}

\paragraph{Agent and harness optimization.}
Agent optimization has expanded from prompts and pre-defined workflows to the executable harness surrounding a model. AFlow searches graphs of model-invoking operators using execution feedback~\citep{zhang2025aflow}. More recent systems directly modify harness components: agentic harness engineering makes edits traceable~\citep{lin2026ahe}, HarnessFix diagnoses failures before applying repairs~\citep{chen2026harnessfix}, and HarnessX composes harnesses from typed primitives using trajectory evidence~\citep{chen2026harnessx}. RHO uses LLMs to evaluate and compare execution trajectories without ground-truth labels, guiding harness evolution toward higher task success~\citep{pan2026rho}. In contrast, our method, \method{}, optimizes both task success rate and monetary cost. Rather than returning a single harness, it produces a set of harnesses with different performance-cost trade-offs, organized into lite, medium, and heavy tiers.

\paragraph{Component optimization and inference-time budget allocation.}
Some methods focus on specific components, such as skills or memory~\citep{pan2026contextpilot}. SkillOpt optimizes skill documents~\citep{yang2026skillopt}, while SkillMOO and FRAMES incorporate cost or regression constraints into skill evolution~\citep{gong2026skillmoo,wang2026frames}. Dynamic Cheatsheet and ReasoningBank retain reusable strategies from past experience~\citep{suzgun2026dynamiccheatsheet,ouyang2026reasoningbank}, while Sleep-time Compute builds reusable memory by reasoning over context before a query arrives~\citep{lin2025sleeptime}. MemoryCPT jointly optimizes memory construction, retrieval, and summarization~\citep{lei2026memorycpt}. Other work addresses budget allocation at inference time: BATS learns a budget-aware policy for allocating model tokens and external tool calls~\citep{liu2025bats}, while COMPAS jointly selects models, prompts, and decoding settings according to task difficulty~\citep{gong2026compas}. \method{} instead uses past trajectories to evolve the entire harness, including instructions, skills, tools, hooks, and verification rules. Table~\ref{tab:method-comparison} summarizes how \method{} differs from prior work.
\begin{table}[t]
\centering
\caption{Comparison of different harness self-improvement methods.}
\label{tab:method-comparison}
\vspace{3pt}
\fontsize{8.5}{10}\selectfont
\definecolor{comparisonHeader}{rgb}{0.94588,0.95764,0.96883}
\definecolor{comparisonOurs}{rgb}{0.92587,0.97171,0.99509}
\setlength{\tabcolsep}{3.14pt}
\renewcommand{\arraystretch}{1.18}
\begin{tabular}{lccccc}
\toprule
\rowcolor{comparisonHeader}
\textbf{Method} & \textbf{Goal} & \textbf{Scope} & \textbf{Feedback} & \textbf{Iteration} & \textbf{Output} \\
\midrule
AFlow \citep{zhang2025aflow} & Pass & Workflow & Ground truth & Recursive & Single \\
SkillMOO \citep{gong2026skillmoo} & Pass + Cost & Skills & Ground truth & Recursive & Pareto set \\
RHO \citep{pan2026rho} & Pass & Skills + tools & Label-free & Single round & Single \\
AHE \citep{lin2026ahe} & Pass & Harness & Ground truth & Recursive & Single \\
\rowcolor{comparisonOurs}
\textbf{\method{}} & Pass + Cost & Harness & Label-free & Recursive & Pareto set \\
\bottomrule
\end{tabular}
\vspace{-3mm}
\end{table}

\section{Problem Setting}
\label{sec:problem-setting}

A \emph{harness} $h\in\mathcal H$ is the software runtime layer that manages a foundation model's interactions and execution. It includes prompts, tools and skills, context and memory mechanisms, hooks, and verification policies~\citep{yehudai2026survey,openai2026agentarchitecture}. Given a task $x$, the harness determines what information the agent receives, which actions it can take, and how those actions are executed, directly influencing both task success and resource consumption.
An agent equipped with harness $h$ solves task $x$, producing a trajectory $\tau_h(x)=\operatorname{Solve}(h,x)$ that records tool calls, environment feedback, and the final artifact. Recursive self-improvement uses trajectories from previous tasks to guide harness evolution and improve performance on future tasks. Below, we define the three evaluation metrics: pass rate, execution cost, and runtime latency.

\paragraph{Pass rate.}
To prevent test-set leakage, we partition tasks into training, validation, and test sets, denoted by $\mathcal D_{\mathrm{train}}$, $\mathcal D_{\mathrm{val}}$, and $\mathcal D_{\mathrm{test}}$. Outcomes on the training set are used to update harnesses, while the validation set is used to evaluate and select candidate harnesses. The test set is used only for final evaluation after evolution. For a task $x\in\mathcal D_{\mathrm{test}}$, the official tests evaluate the final artifact produced by $\tau_h(x)$. We set $\operatorname{Pass}(\tau_h(x))=1$ if all official tests pass and $\operatorname{Pass}(\tau_h(x))=0$ otherwise. The pass rate is the fraction of test tasks that pass all official tests, $\mathrm{PR}(h)=\frac{1}{|\mathcal D_{\mathrm{test}}|}\sum_{x\in\mathcal D_{\mathrm{test}}}\operatorname{Pass}\!\left(\tau_h(x)\right)$.

\paragraph{Cost.}
We compute execution cost from token usage, distinguishing uncached input, cached input, and output tokens because they are billed at different prices. For each task $x$, we aggregate these token counts over all model calls in $\tau_h(x)$ and compute its cost $c_h(x)$ using the prices in Table~\ref{tab:token-pricing} (Appendix~\ref{apx:implement-details}). We report the average cost per test task as $C(h)=\frac{1}{|\mathcal D_{\mathrm{test}}|}\sum_{x\in\mathcal D_{\mathrm{test}}}c_h(x)$.

\paragraph{Latency.}
We define task latency as the wall-clock time from the start of agent execution to its completion: $\ell_h(x)=T_{\mathrm{end}}(\tau_h(x))-T_{\mathrm{start}}(\tau_h(x))$.
The time spent grading the final output is not included in task latency. We average task latencies over the test set to obtain the mean latency, $L(h)=\frac{1}{|\mathcal D_{\mathrm{test}}|}\sum_{x\in\mathcal D_{\mathrm{test}}}\ell_h(x)$. Because model response times fluctuate with server load and network conditions, we treat latency as an auxiliary metric rather than an optimization objective.

\paragraph{Optimization objective.}
We aim to evolve harnesses that improve performance while reducing cost. We formulate the optimization problem as
\begin{equation}
    \min_{h\in\mathcal H}\left(-\mathrm{PR}(h),\,C(h)\right)\quad\text{s.t.}\quad\mathrm{PR}(h)\geq\mathrm{PR}(h_0).
    \label{eq:optimization-objective}
\end{equation}
Here, $h_0$ is the initial harness. We seek the \textbf{\textit{Pareto}} frontier and select three \textbf{\textit{representative}} operating points: cost-focused (\emph{lite}), balanced (\emph{medium}), and performance-focused (\emph{heavy}). However, finding these trade-offs poses two challenges. \textbf{(1) A vast harness search space.} A harness comprises components such as instructions, tools, skills, hooks, and rules, all of which can be modified during evolution. Their possible combinations create a vast, discrete search space that cannot be optimized using gradients. \textbf{(2) Balancing performance and cost.} Additional reasoning, exploration, or verification can improve performance but also consumes more tokens, increasing cost. Conversely, indiscriminately reducing these activities may hurt performance. Advancing the Pareto frontier thus requires analyzing execution trajectories alongside task outcomes and token usage to distinguish necessary computation from avoidable overhead and find the ``sweet spot'' that reduces cost without compromising performance.

\section{HarnessPro}
\label{sec:method}

\begin{figure}[t]
    \centering
    \includegraphics[width=\linewidth]{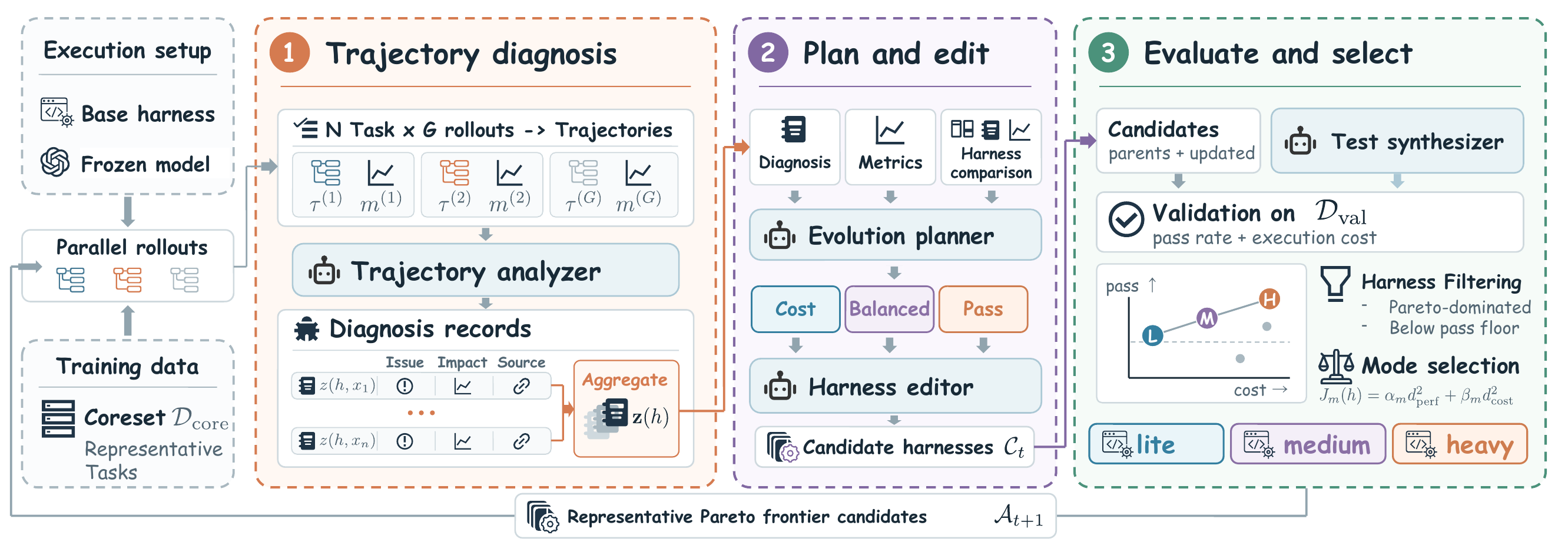}
    \vspace{-5mm}
    \caption{Overview of \method{}. In each epoch, it analyzes rollout trajectories, generates candidate harnesses, and selects representatives based on performance and cost.}
    \label{fig:harnesspro-overview}
\end{figure}

To address these challenges, we introduce \method{}, a multi-agent framework that analyzes execution trajectories and iteratively modifies harnesses to advance the performance-cost Pareto frontier. We first outline the evolution process, then describe trajectory-based diagnosis and candidate generation, followed by test-based evaluation and selection.

\subsection{Overview}
\label{sec:method-overview}

To keep trajectory collection and analysis manageable, we begin by selecting a training coreset $\mathcal D_{\mathrm{train}}$. Following RHO~\citep{pan2026rho}, we use the Determinantal Point Process (DPP) kernel~\citep{kulesza2012dpp} to balance task difficulty and diversity (Appendix~\ref{apx:dpp-coreset}). Starting from an executable harness $h_0$, we maintain a \textbf{\textit{candidate set}} $\mathcal A_t$ of harnesses at epoch $t$, initialized as $\mathcal A_0=\{h_0\}$. In each epoch, \method{} employs a multi-agent framework to analyze trajectories of these candidate harnesses on the coreset $\mathcal D_{\mathrm{train}}$ to identify task failures, unproductive exploration, and redundant computation. The framework then plans and implements harness modifications to address these issues, producing new executable candidate harnesses as described in Section~\ref{sec:diagnosis-generation}.

We evolve harnesses in a label-free setting, \textbf{\textit{without}} access to ground-truth solutions, official test cases, or graders. Instead of using an LLM to assess and compare trajectories directly, as in RHO~\citep{pan2026rho}, we use it to generate tests from task descriptions and trajectories produced by the initial harness. In each epoch, we evaluate candidate harnesses on the validation set using the generated tests, measuring \textbf{\textit{pass rate}} and \textbf{\textit{execution cost}}. After discarding candidates that fall below the performance floor or are Pareto-dominated, we use normalized weighted squared-distance selection to retain \textbf{\textit{representatives}} for cost-focused (lite), balanced (medium), and performance-focused (heavy) modes. The selected harnesses form the candidate set for the next epoch. This process repeats until the epoch limit is reached, after which we fix the harnesses and evaluate them on the test set. Algorithm~\ref{alg:harnesspro} summarizes the full procedure.

\begin{algorithm}[t]
\caption{Cost-aware recursive harness self-improvement with \method{}.}
\label{alg:harnesspro}
\footnotesize
\definecolor{harnessStageBlue}{rgb}{0.18431,0.30980,0.43529}
\newcommand{\harnessstage}[1]{\Statex\hspace{\algorithmicindent}\textcolor{harnessStageBlue}{\rule[1pt]{6pt}{6pt}\,\textbf{#1}}}
\algrenewcommand\algorithmicrequire{\textbf{Input:}}
\algrenewcommand\algorithmicensure{\textbf{Output:}}
\begin{algorithmic}[1]
\Require Seed harness $h_0$; data $\mathcal D_{\mathrm{train}},\mathcal D_{\mathrm{val}}$; prompts $I_1$--$I_3$; rollouts $G$; epoch limit $T$
\Ensure Harnesses for the lite $h^{T}_{\mathrm{lite}}$, medium $h^{T}_{\mathrm{medium}}$, and heavy $h^{T}_{\mathrm{heavy}}$ modes.
\State Collect $\tau_{h_0}(x)$ for each $x\in\mathcal D_{\mathrm{train}}\cup\mathcal D_{\mathrm{val}}$
\State $\mathcal S(x)\gets\operatorname{\mathbf{TestSynth}}(x,\tau_{h_0}(x))$ for each $x$ \Comment{Generate once}
\State Evaluate $h_0$ on $\mathcal D_{\mathrm{val}}$ using $\mathcal S$; fix its initial pass rate $\mathrm{PR}_{\mathrm{val}}(h_0)$ as the performance floor
\State $\mathcal A_0\gets\{h_0\}$; $t\gets0$; $h_m^0\gets h_0$ for $m\in\{\mathrm{lite},\mathrm{medium},\mathrm{heavy}\}$
\While{$t<T$}
    \harnessstage{Stage 1\quad Trajectory diagnosis}
    \For{each $h\in\mathcal A_t$}
        \State Collect $G$ independent rollouts per $x\in\mathcal D_{\mathrm{train}}$, recording $\tau_h^{t,g}(x)$ and $m_h^{t,g}(x)$ using $\mathcal S(x)$
        \State $z(h,x)\gets\operatorname{\mathbf{Analyzer}}(x,\{(\tau_h^{t,g}(x),m_h^{t,g}(x))\}_{g=1}^{G})$ for each $x\in\mathcal D_{\mathrm{train}}$
        \State $\mathbf z(h)\gets\bigcup\nolimits_{x\in\mathcal D_{\mathrm{train}}}z(h,x)$; $\mathbf m(h)\gets\bigcup\nolimits_{x\in\mathcal D_{\mathrm{train}}}\{m_h^{t,g}(x)\}_{g=1}^{G}$
    \EndFor
    \harnessstage{Stage 2\quad Candidate generation}
    \For{each $h_i^t\in\mathcal A_t$ and $j\in\{1,2,3\}$ \textbf{in parallel}}
        \State Generate a plan from $I_j$, diagnoses, metrics, and cross-candidate evidence:
        \State $\mathbf p_j(h_i^t)\gets\operatorname{\mathbf{Plan}}(I_j,h_i^t,\mathbf z(h_i^t),\mathbf m(h_i^t),\{(\operatorname{Diff}(h_i^t,h_k^t),\mathbf z(h_k^t),\mathbf m(h_k^t))\}_{k\ne i})$
        \State $\widetilde h_{i,j}^t\gets\operatorname{\mathbf{Edit}}(h_i^t,\mathbf z(h_i^t),\mathbf p_j(h_i^t))$; commit $\widetilde h_{i,j}^t$ to Git \Comment{Separate workspace}
    \EndFor
    \State $\mathcal C_t\gets\{\widetilde h_{i,j}^t\}_{i,j}$; $\widetilde{\mathcal C}_t\gets\mathcal A_t\cup\mathcal C_t$
    \harnessstage{Stage 3\quad Evaluation and selection}
    \State Evaluate each $h\in\widetilde{\mathcal C}_t$ once per $x\in\mathcal D_{\mathrm{val}}$ using the fixed $\mathcal S(x)$; aggregate $\mathbf m_{\mathrm{val}}(h)$
    \State Filter candidates: $\widetilde{\mathcal A}_{t+1}\gets\{h\in\widetilde{\mathcal C}_t\mid\mathrm{PR}_{\mathrm{val}}(h)\geq\mathrm{PR}_{\mathrm{val}}(h_0)\;\land\;\not\exists h'\in\widetilde{\mathcal C}_t:h'\prec h\}$
    \State Select representatives: $h_m^{t+1}\gets\operatorname*{arg\,min}_{h\in\widetilde{\mathcal A}_{t+1}}J_m(h)$ for $m\in\{\mathrm{lite},\mathrm{medium},\mathrm{heavy}\}$ \Comment{Eq.~\ref{eq:mode-selection}}
    \State Update candidates: $\mathcal A_{t+1}\gets\{h_{\mathrm{lite}}^{t+1},h_{\mathrm{medium}}^{t+1},h_{\mathrm{heavy}}^{t+1}\}$; $t\gets t+1$
\EndWhile
\State \Return $(h_{\mathrm{lite}}^T,h_{\mathrm{medium}}^T,h_{\mathrm{heavy}}^T)$ \Comment{Fixed before held-out test evaluation}
\end{algorithmic}
\end{algorithm}

\subsection{Trajectory Analysis and Harness Evolution}
\label{sec:diagnosis-generation}

At epoch $t$, we run each harness in the current candidate set $\mathcal A_t=\{h_i^t\}_{i=1}^{|\mathcal A_t|}$ independently $G$ times on every training task $x\in\mathcal D_{\mathrm{train}}$. For candidate harness $h\in\mathcal A_t$, this rollout process yields a trajectory $\tau_h^{t,g}(x)$ and a metric vector $m_h^{t,g}(x)=\left(\operatorname{Pass}_h^{t,g}(x),\ c_h^{t,g}(x),\ \ell_h^{t,g}(x)\right)$. Here, $\operatorname{Pass}_h^{t,g}(x)$ is 1 if the artifact produced in $\tau_h^{t,g}(x)$ passes all generated tests in $\mathcal S(x)$, while $c_h^{t,g}(x)$ and $\ell_h^{t,g}(x)$ record its execution cost and latency. We also retain token usage and tool-call details for later diagnosis. These trajectories and artifacts guide harness self-improvement through a multi-agent framework comprising a \textbf{\textit{trajectory analyzer}}, an \textbf{\textit{evolution planner}}, a \textbf{\textit{harness editor}}, and a \textbf{\textit{test synthesizer}}. We describe each agent below.

\paragraph{Trajectory Analyzer.}
Execution trajectories can reveal harness weaknesses, but the relevant evidence is often dispersed across long sequences of model responses and tool outputs, making these weaknesses difficult to locate and interpret. We therefore implement a trajectory-analysis agent to diagnose these problems for each candidate $h$ and task $x$ at epoch $t$. To reduce the impact of randomness, the agent jointly examines the trajectories $\{\tau_h^{t,g}(x)\}_{g=1}^{G}$ from $G$ independent rollouts and their metrics $\{m_h^{t,g}(x)\}_{g=1}^{G}$, rather than relying on a single attempt:
\begin{equation}
    z(h,x)
    = 
    \operatorname{Analyzer}\!\left(
    x,
    \left\{\left(\tau_h^{t,g}(x),m_h^{t,g}(x)\right)\right\}_{g=1}^{G}
    \right).
    \label{eq:trajectory-diagnoses}
\end{equation}
The analyzer's output $z(h,x)$ reports (1) identified problems, (2) their impact on performance and cost, and (3) their sources and supporting evidence. The evidence includes a concise trajectory summary or relevant excerpts, together with the path of the trajectory file and the relevant event range. Two examples are provided in Appendix~\ref{app:diagnosis-example}. We combine diagnoses across different tasks as $\mathbf z(h)=\bigcup\nolimits_{x\in\mathcal D_{\mathrm{train}}}z(h,x)$. If $\mathbf z(h)$ exceeds 64K tokens, the analyzer summarizes it. We similarly collect the metric vectors across training tasks as $\mathbf m(h)=\bigcup\nolimits_{x\in\mathcal D_{\mathrm{train}}}\{m_h^{t,g}(x)\}_{g=1}^{G}$. The planner uses these diagnoses and metrics to guide its proposals.

\paragraph{Evolution Planner.}
Building on these diagnoses, the planner proposes three improvement plans for each candidate. Each plan is guided by a distinct instruction: $I_1$ emphasizes cost reduction, $I_2$ balances performance and cost, and $I_3$ prioritizes performance, as detailed in Appendix~\ref{app:evolution-prompt}. The planner starts with the full implementation of $h_i^t$, using its diagnoses and metrics to identify where modifications are needed. It also compares harness $h_i^t$ and the other candidates $h_k^t$ by examining their code differences $\operatorname{Diff}(h_i^t,h_k^t)$:
\begin{equation}
\mathbf p_j(h_i^t) = \operatorname{Plan}(
    I_j,h_i^t,\mathbf z(h_i^t),\mathbf m(h_i^t),\{(
    \operatorname{Diff}(h_i^t,h_k^t),
    \mathbf z(h_k^t),
    \mathbf m(h_k^t)
    )\}_{\substack{k\neq i}}
    ).
\label{eq:evolution-plan}
\end{equation}
By comparing candidates, the planner can identify effective mechanisms and remaining bottlenecks. Each plan explains which components to modify, how to modify them, and the expected effects on performance and cost, as exemplified in Appendix~\ref{app:planner-examples}. Full trajectories are not provided to the planner by default. When needed, it can inspect them using the file paths recorded in the diagnoses. The planner is instructed to propose generalizable, rather than task-specific, harness updates. Throughout evolution, the validation and test sets remain hidden from the planner to avoid benchmark leakage.

\paragraph{Harness Editor.}
We then design a harness editor to modify candidate $h_i^t$, producing a new, executable candidate harness using the proposed plan $\mathbf p_j(h_i^t)$ and the diagnoses $\mathbf z(h_i^t)$:
\begin{equation}
    \widetilde h_{i,j}^t
    = \operatorname{Edit}\!\left(
    h_i^t,\mathbf z(h_i^t),\mathbf p_j(h_i^t)
    \right),
    \qquad j=1,2,3.
    \label{eq:harness-edit}
\end{equation}
Each of the three plans is implemented in a separate workspace, and the resulting harness is committed to Git so that changes can be traced and reverted. Newly generated harnesses form $\mathcal C_t$, and we combine them with the parent harnesses in $\mathcal A_t$ for evaluation:
\begin{equation}
    \mathcal C_t
    = \bigcup_{i=1}^{|\mathcal A_t|}
    \left\{\widetilde h_{i,1}^t,\widetilde h_{i,2}^t,\widetilde h_{i,3}^t\right\},
    \qquad
    \widetilde{\mathcal C}_t = \mathcal A_t\cup\mathcal C_t.
    \label{eq:candidate-pool}
\end{equation}
All harnesses in $\widetilde{\mathcal C}_t$ then undergo the same evaluation and selection process, as described below.

\subsection{Evaluation and Candidate Selection}
\label{sec:evaluation-selection}

\begin{wrapfigure}{r}{0.40\textwidth}
    \vspace{-\intextsep}
    \centering
    \setlength{\abovecaptionskip}{2pt}
    \setlength{\belowcaptionskip}{4pt}
    \includegraphics[width=\linewidth]{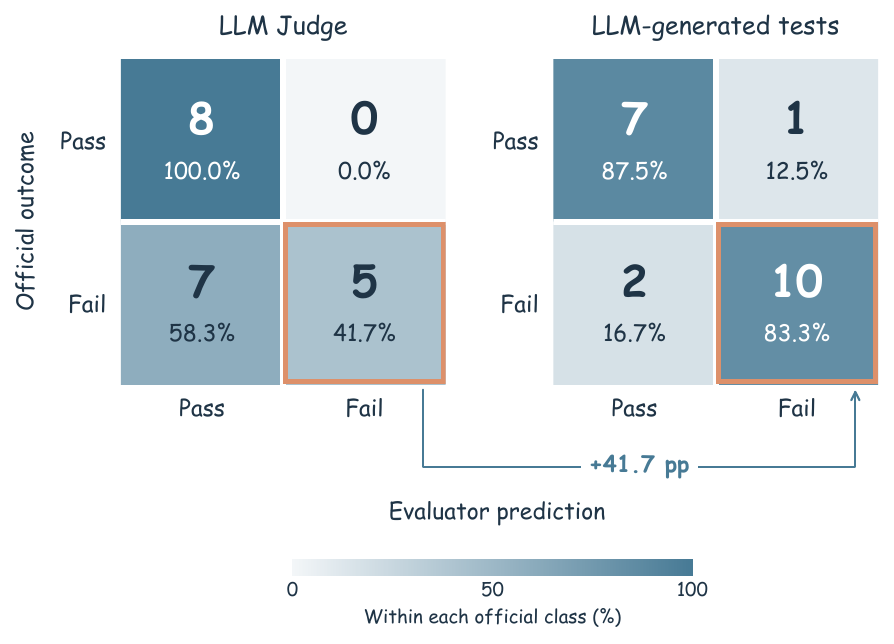}
    \caption{Confusion matrices pooled across training tasks from Terminal-Bench~2 and SWE-Bench Pro.}
    \label{fig:judge-test-agreement}
\end{wrapfigure}
\paragraph{Test Synthesizer.}
During harness evolution, we follow RHO~\citep{pan2026rho}, adopting a label-free setting without access to ground-truth solutions, official tests, or grader feedback.
RHO relies on LLM self-preference to select harnesses.
However, we find that LLM judges tend to be overly optimistic when assessing the execution trajectories, often treating flawed implementations as successful. We therefore introduce a dedicated test-synthesis agent to generate test cases for evaluation; its instruction, skill, and validation tool are outlined in Appendix~\ref{app:test-synthesizer-setup}.
For each task $x\in\mathcal D_{\mathrm{train}}\cup\mathcal D_{\mathrm{val}}$, the agent generates tests from its description $x$ and the initial trajectory $\tau_{h_0}(x)$:
\begin{equation}
    \mathcal S(x)=\operatorname{TestSynth}\!\left(x,\tau_{h_0}(x)\right).
    \label{eq:test-synthesis}
\end{equation}
These generated tests remain fixed throughout the self-improvement process, ensuring a fair comparison among candidate harnesses. 
A preliminary experiment demonstrates the effectiveness of this approach, with failure recall increasing from 41.7\% to 83.3\% compared with direct LLM judgments (Figure~\ref{fig:judge-test-agreement}). 
After each epoch, we run each candidate $h\in\widetilde{\mathcal C}_t$ once per validation task $x$ and check its output against $\mathcal S(x)$. Task-level pass indicators, costs, and latencies are averaged to obtain the validation metrics $\mathbf m_{\mathrm{val}}(h)=\left(\mathrm{PR}_{\mathrm{val}}(h),C_{\mathrm{val}}(h),L_{\mathrm{val}}(h)\right)$.

\paragraph{Candidate filtering and selection.}
To prevent the candidate set from growing indefinitely, we retain at most three harnesses for the next epoch. We first discard candidates below the initial pass-rate floor $\mathrm{PR}_{\mathrm{val}}(h_0)$ or Pareto-dominated by another candidate:
\begin{equation}
    \widetilde{\mathcal A}_{t+1}
    = \left\{h\in\widetilde{\mathcal C}_t\ \middle|\
    \mathrm{PR}_{\mathrm{val}}(h)
    \geq \mathrm{PR}_{\mathrm{val}}(h_0)
    \;\land\;
    \not\exists h'\in\widetilde{\mathcal C}_t:h'\prec h
    \right\}.
    \label{eq:pareto-filter}
\end{equation}
\begin{wrapfigure}{r}{0.40\textwidth}
    \vspace{-\intextsep}
    \centering
    \setlength{\abovecaptionskip}{2pt}
    \setlength{\belowcaptionskip}{-5pt}
    \includegraphics[width=\linewidth]{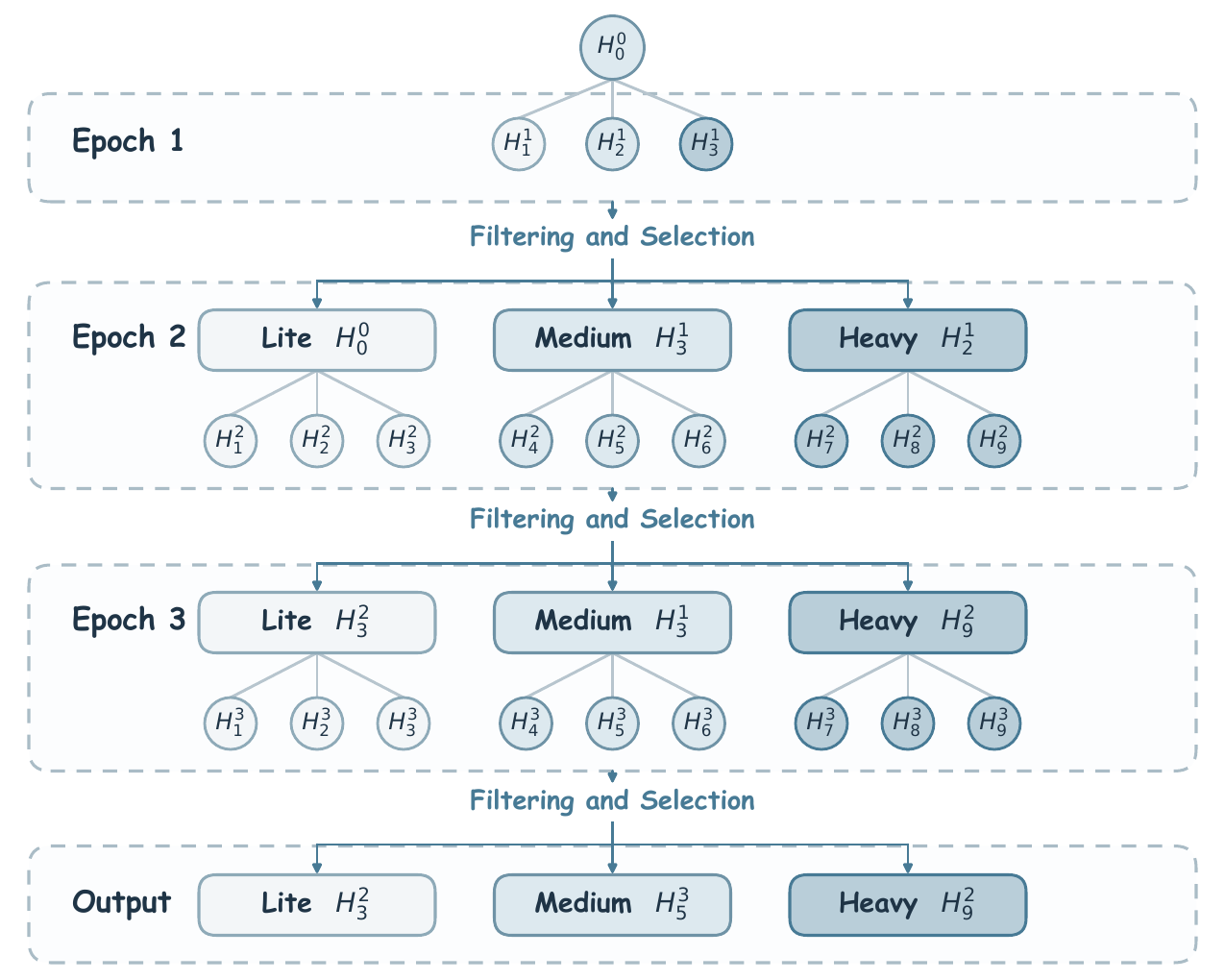}
    \caption[Schematic of recursive harness update, filtering, and selection.]{Schematic of recursive harness update, filtering, and selection.\protect\footnotemark}
    \label{fig:recursive-harness-selection}
\end{wrapfigure}
\footnotetext{In practice, filtering and selection may retain fewer than three distinct harnesses, see Figure~\ref{fig:evolution-pareto}.}
\noindent Here, $h'\prec h$ denotes Pareto dominance: $h'$ is no worse than $h$ on both metrics and strictly better on at least one. We select representatives using a normalized weighted squared-distance criterion~\citep{marler2004survey} that accounts for actual differences in performance and cost. We first measure each candidate's normalized deviations from the highest pass rate and lowest cost:
\begin{equation}
    \begin{aligned}
        d_{\mathrm{perf}}(h)
        &= (P_{\max}-\mathrm{PR}_{\mathrm{val}}(h))/(P_{\max}-P_{\min}),\\
        d_{\mathrm{cost}}(h)
        &= (C_{\mathrm{val}}(h)-C_{\min})/(C_{\max}-C_{\min}).
    \end{aligned}
    \label{eq:normalized-losses}
\end{equation}
Here, $P_{\min},P_{\max}$ and $C_{\min},C_{\max}$ are the extrema of validation pass rate and mean cost over the current filtered set $\widetilde{\mathcal A}_{t+1}$. 
The selection objective for mode $m$ is $J_m(h)=\alpha_m d_{\mathrm{perf}}(h)^2+\beta_m d_{\mathrm{cost}}(h)^2$, where $\alpha_m$ and $\beta_m$ weight performance and cost, respectively. Each mode $m\in\{\mathrm{lite},\mathrm{medium},\mathrm{heavy}\}$ independently selects the candidate with the smallest weighted squared distance:
\begin{equation}
    h_m^{t+1}
    = \operatorname{arg\,min}_{h\in\widetilde{\mathcal A}_{t+1}} J_m(h).
    \label{eq:mode-selection}
\end{equation}
The selected harnesses form $\mathcal A_{t+1}$ for the next round, as illustrated in Figure~\ref{fig:recursive-harness-selection}. Filtering and selection use only validation metrics, with no access to test-set results. The final harnesses are evaluated on the held-out test set $\mathcal D_{\mathrm{test}}$ only after all rounds of recursive self-improvement are complete.

\section{Experiments}
\label{sec:experiments}

\subsection{Experimental Setup}
\label{sec:experimental-setup}

\paragraph{Data and implementation.}
We evaluate \method{} on Terminal-Bench~2~\citep{tbench2025}, SWE-Bench Pro~\citep{deng2026swebenchpro}, and ProgramBench~\citep{yang2026programbench}, covering terminal tasks, repository-level issue resolution, and program reconstruction.
Following the dataset splitting protocol of RHO~\citep{pan2026rho}, we hold out 59, 100, and 100 tasks, respectively, for testing.
From the remaining tasks in each benchmark, we select a 10-task training set $\mathcal D_{\mathrm{train}}$ using the DPP-based coreset selection method~\citep{kulesza2012dpp} described in Section~\ref{sec:method-overview}.
The remaining tasks form the validation set.
We pair GPT-5.5 with \texttt{Codex 0.147.0} and GLM-5.2 with \texttt{Claude Code 2.1.231}.
Each candidate harness produces $G=3$ rollouts per training task in each epoch.
Evolution runs for three epochs on Terminal-Bench~2 and SWE-Bench Pro, and four on ProgramBench, retaining up to three harnesses after each epoch.
All methods use the same data splits and grading settings. Further implementation details are in Appendix~\ref{apx:implement-details}.

\paragraph{Baselines.}
We compare \method{} against several strong harness optimization methods.
The original Codex or Claude Code harness is \emph{Vanilla}, the starting point for optimization.
\emph{Dynamic Cheatsheet}~\citep{suzgun2026dynamiccheatsheet} maintains reusable strategies and code snippets, \emph{ReasoningBank}~\citep{ouyang2026reasoningbank} retrieves strategies distilled from past trajectories, and \emph{Sleep-time Compute}~\citep{lin2025sleeptime} prepares reusable memory before new tasks.
\emph{RHO}~\citep{pan2026rho} revises harness components through trajectory analysis and selects candidates using LLM self-preference.
To separate the effects of cost-aware and recursive optimization, we include two RHO variants.
\emph{Balance} changes only RHO's optimization instruction, asking for better task performance at lower cost.
\emph{RHO+Iteration} runs \emph{Balance} for the same number of epochs as \method{}.
Each subsequent epoch starts from the harness selected in the preceding epoch by the best-of-$N$ procedure of RHO.

\paragraph{Evaluation metrics.} On Terminal-Bench~2 and SWE-Bench Pro, a task passes only if all its official tests pass.
For ProgramBench, we follow its official evaluation protocol~\citep{yang2026programbench} and report pass rate as the macro-average of the fraction of test cases passed for each task.
We also report mean cost $C(h)$ and latency $L(h)$. For \method{}, all metrics are given as mean $\pm$ sample standard deviation over three runs on Terminal-Bench~2 and SWE-Bench Pro, and two on ProgramBench.
The token prices used to calculate cost are listed in Table~\ref{tab:token-pricing} in the appendix. 

\subsection{Main Results}
\label{sec:main-results}

\newcommand{\resultmsd}[2]{\mbox{#1{\scriptsize$\pm\,#2$}}}
\begin{table}[htbp]
\centering
\caption{Performance and efficiency across three benchmarks.
P denotes pass rate (\%), while C and T denote mean cost (\$) and execution time (s).
We report mean and standard deviation for \method{}.
On ProgramBench, P is the macro-average of the fraction of tests passed per task.}
\label{tab:main-results}
\vspace{4pt}
\small
\definecolor{harnessTableHeader}{rgb}{0.94588,0.95764,0.96883}
\definecolor{harnessTableOurs}{rgb}{0.92587,0.97171,0.99509}
\setlength{\dashlinedash}{2pt}
\setlength{\dashlinegap}{1.5pt}
\renewcommand{\arraystretch}{1.18}
\setlength{\tabcolsep}{3pt}
\resizebox{\linewidth}{!}{%
\begin{tabular}{@{}l*{9}{c}@{}}
\toprule
\rowcolor{harnessTableHeader}
& \multicolumn{3}{c}{\textbf{Terminal-Bench 2}}
& \multicolumn{3}{c}{\textbf{SWE-Bench Pro}}
& \multicolumn{3}{c}{\textbf{ProgramBench}} \\
\rowcolor{harnessTableHeader}
\multirow{-2}{*}{\textbf{Method}}
& \textbf{P(\%)}$\;\uparrow$ & \textbf{C(\$)}$\;\downarrow$ & \textbf{T(s)}$\;\downarrow$
& \textbf{P(\%)}$\;\uparrow$ & \textbf{C(\$)}$\;\downarrow$ & \textbf{T(s)}$\;\downarrow$
& \textbf{P(\%)}$\;\uparrow$ & \textbf{C(\$)}$\;\downarrow$ & \textbf{T(s)}$\;\downarrow$ \\
\specialrule{\lightrulewidth}{0pt}{0pt}
\multicolumn{10}{c}{\textbf{GPT-5.5 $\times$ Codex}} \\
\specialrule{\lightrulewidth}{0pt}{0pt}
Vanilla & 69.5 & 0.89 & 260 & 59.0 & 1.88 & 337 & 60.3 & 2.65 & 669 \\
\hdashline
Dynamic Cheatsheet & 72.9 & 1.06 & 421 & 62.0 & 2.19 & 335 & 68.5 & 3.58 & 742 \\
ReasoningBank & 74.6 & 0.99 & 369 & 61.0 & 2.13 & 341 & 62.5 & 3.06 & 661 \\
Sleep-time Compute & 74.6 & 1.04 & 331 & 62.0 & 2.26 & 337 & 67.7 & 3.38 & \textbf{637} \\
RHO & 76.3 & 1.09 & 354 & 80.0 & 2.68 & 437 & 68.8 & 3.59 & 928 \\
Balance & 74.6 & 0.79 & 355 & 78.0 & 2.08 & 399 & 65.7 & 3.29 & 865 \\
RHO+Iteration & 76.3 & 0.82 & 328 & 80.0 & 1.90 & 375 & 68.3 & 3.13 & 789 \\
\hdashline
\rowcolor{harnessTableOurs}
\textbf{HarnessPro-Lite} & \resultmsd{72.9}{1.7} & \resultmsd{\textbf{0.43}}{0.01} & \resultmsd{\textbf{230}}{32} & \resultmsd{72.3}{0.6} & \resultmsd{\textbf{0.80}}{0.02} & \resultmsd{\textbf{203}}{19} & \resultmsd{64.3}{0.04} & \resultmsd{\textbf{2.18}}{0.06} & \resultmsd{985}{70} \\
\rowcolor{harnessTableOurs}
\textbf{HarnessPro-Medium} & \resultmsd{76.3}{1.7} & \resultmsd{0.56}{0.04} & \resultmsd{253}{35} & \resultmsd{79.0}{1.0} & \resultmsd{1.26}{0.03} & \resultmsd{326}{25} & \resultmsd{67.9}{0.47} & \resultmsd{2.90}{0.13} & \resultmsd{745}{61} \\
\rowcolor{harnessTableOurs}
\textbf{HarnessPro-Heavy} & \resultmsd{\textbf{80.8}}{1.0} & \resultmsd{0.72}{0.05} & \resultmsd{310}{31} & \resultmsd{\textbf{80.7}}{0.6} & \resultmsd{1.97}{0.02} & \resultmsd{333}{10} & \resultmsd{\textbf{69.6}}{1.03} & \resultmsd{2.95}{0.13} & \resultmsd{888}{32} \\
\specialrule{\lightrulewidth}{0pt}{0pt}
\multicolumn{10}{c}{\textbf{GLM-5.2 $\times$ Claude Code}} \\
\specialrule{\lightrulewidth}{0pt}{0pt}
Vanilla & 74.6 & 0.92 & 963 & 56.0 & 1.17 & 402 & 59.7 & 11.34 & \textbf{3847} \\
RHO & 78.0 & 1.00 & 1318 & 75.0 & 2.17 & 520 & 63.3 & 17.00 & 6434 \\
\hdashline
\rowcolor{harnessTableOurs}
\textbf{HarnessPro-Lite} & \resultmsd{75.1}{1.0} & \resultmsd{\textbf{0.43}}{0.03} & \resultmsd{\textbf{665}}{8} & \resultmsd{70.7}{0.6} & \resultmsd{\textbf{0.70}}{0.02} & \resultmsd{\textbf{289}}{3} & \resultmsd{60.1}{0.6} & \resultmsd{\textbf{11.10}}{0.28} & \resultmsd{3907}{32} \\
\rowcolor{harnessTableOurs}
\textbf{HarnessPro-Medium} & \resultmsd{76.3}{0.0} & \resultmsd{0.50}{0.01} & \resultmsd{968}{17} & \resultmsd{75.0}{1.0} & \resultmsd{0.87}{0.01} & \resultmsd{332}{10} & \resultmsd{63.0}{0.4} & \resultmsd{13.74}{0.16} & \resultmsd{5893}{132} \\
\rowcolor{harnessTableOurs}
\textbf{HarnessPro-Heavy} & \resultmsd{\textbf{81.4}}{1.7} & \resultmsd{0.62}{0.02} & \resultmsd{1080}{15} & \resultmsd{\textbf{77.3}}{0.6} & \resultmsd{1.01}{0.03} & \resultmsd{412}{11} & \resultmsd{\textbf{64.4}}{0.8} & \resultmsd{14.95}{0.08} & \resultmsd{5925}{110} \\
\bottomrule
\end{tabular}%
}
\end{table}

Table~\ref{tab:main-results} compares \method{} with several baseline methods across three benchmarks.
With GPT-5.5 and Codex, RHO improves performance over Vanilla on all three benchmarks, but also increases execution cost.
On SWE-Bench Pro, for example, its mean cost per task rises by 42.6\%, from \$1.88 to \$2.68.
In contrast, the heavy mode of \method{} achieves the \textbf{\textit{highest}} pass rate across all three benchmarks \textbf{\textit{at lower cost than RHO}}.
On Terminal-Bench~2, the heavy mode of \method{} reaches a mean pass rate of 80.8\% across three runs, 4.5 percentage points above RHO, with 33.9\% lower mean cost.
The advantage extends to ProgramBench, where its \textbf{69.6\%} pass rate exceeds RHO's 68.8\% at \textbf{17.8\%} lower mean cost (\$2.95 vs.\ \$3.59).
These results show that better task performance need not come at a higher execution cost.
For the medium mode, we retain strong task performance while reducing cost further.
On SWE-Bench Pro, it achieves a 79.0\% pass rate compared with RHO's 80.0\%, while reducing the average cost per task by 53.0\%, from \$2.68 to \$1.26.
The lite mode lowers cost further, outperforming Vanilla on all three benchmarks while \textbf{spending 17.7--57.4\% less}, making it an economical choice for deployment under strict budget constraints.
Even with comparable iteration budgets, \method{} achieves a better performance--cost trade-off than \textit{RHO+Iteration}. Our medium mode matches the performance of \textit{RHO+Iteration} on Terminal-Bench~2 at 31.7\% lower cost. Moreover, on ProgramBench, our heavy mode raises pass rate by 1.3 percentage points while lowering mean cost by 5.8\%.

\noindent\begin{minipage}{\linewidth}
The results with GLM-5.2 and Claude Code follow a similar pattern: the heavy mode of \method{} surpasses RHO on all three benchmarks at lower cost, while the medium and lite modes outperform Vanilla in pass rate, often at \textbf{\textit{substantially lower cost}}.
Moreover, although latency is not an objective, cost optimization yields faster task completion in most cases.
For example, on SWE-Bench Pro, our heavy mode \textbf{\textit{reduces mean latency}} from RHO's 520 to 412 seconds.
\end{minipage}

\subsection{Analysis of Evolved Harnesses}
\label{sec:evolved-harness-analysis}

To understand how \method{} improves the performance-cost trade-off, we inspect its evolved harnesses.
On Terminal-Bench~2 and SWE-Bench Pro, they select an execution profile from each task description.
Each profile specifies a reasoning effort level and a tool-call budget.
The lite, medium, and heavy modes use \textbf{different rubrics} for profile selection.
Runtime hooks \textbf{enforce} the selected budget.
For example, on Terminal-Bench~2, when \method{}-Medium selects the standard profile, the hooks remind the agent to produce parallel tool calls and avoid redundant checks after 14 tool calls, focus on completing and verifying the task after 26, and block further calls after 44.
The reminders are appended to tool results, preserving the prompt prefix for cache reuse. 
Alongside these dynamic mechanisms, the evolved harnesses also reduce \textbf{static context redundancy}.
On Terminal-Bench~2, for instance, tool-output \textbf{compression} like RTK~\citep{szymkowiak2026rtk} folds repeated logs and condenses long outputs, while \textbf{streamlined} developer instructions reduce redundant input.
This combination curbs recurring prompt overhead and limits the context growth caused by verbose tool feedback, thereby reducing execution cost and latency.
Implementation details of these harnesses are in Appendix~\ref{app:tb2-artifacts}.

\subsection{Evolution Dynamics}
\label{sec:evolution-dynamics}

\begin{figure}[t]
\centering
\begin{minipage}[t]{0.325\linewidth}
  \centering
  \includegraphics[width=\linewidth]{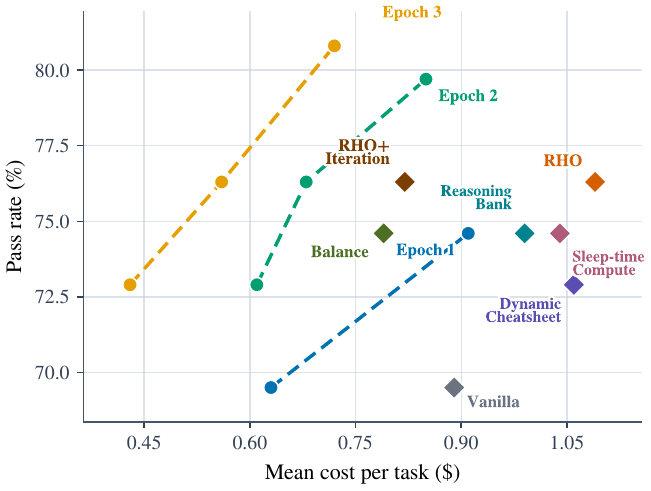}
  \par\smallskip
  {\footnotesize (a) Terminal-Bench 2}
\end{minipage}\hfill
\begin{minipage}[t]{0.325\linewidth}
  \centering
  \includegraphics[width=\linewidth]{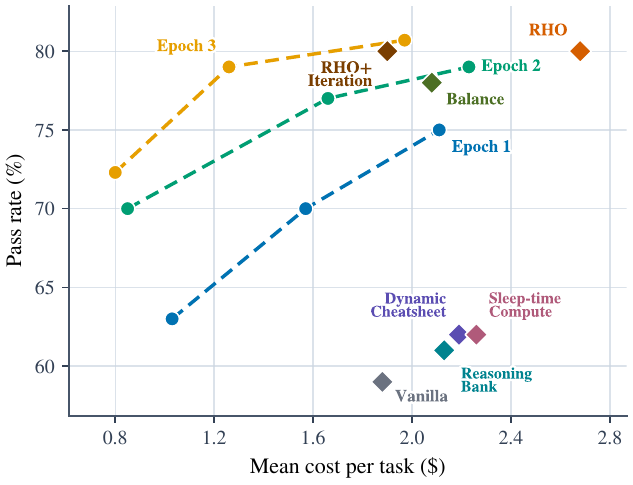}
  \par\smallskip
  {\footnotesize (b) SWE-Bench Pro}
\end{minipage}\hfill
\begin{minipage}[t]{0.325\linewidth}
  \centering
  \includegraphics[width=\linewidth]{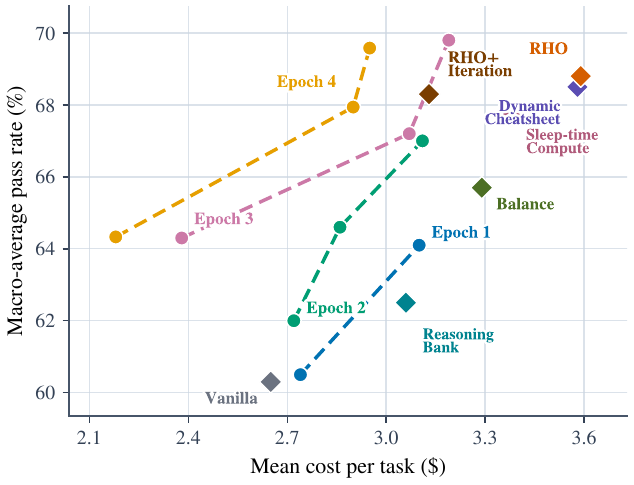}
  \par\smallskip
  {\footnotesize (c) ProgramBench}
\end{minipage}
\caption{Evolution dynamics of \method{} with GPT-5.5 and Codex. Dashed lines connect the harnesses retained in each epoch, and diamonds denote baselines. Intermediate-epoch test-set metrics are reported for visualization only and were not used for filtering or selection. Final-epoch points show means across repeated test runs.}
\label{fig:evolution-pareto}
\end{figure}

Figure~\ref{fig:evolution-pareto} tracks test-set performance and cost across epochs. 
Across all three benchmarks, \method{} advances the performance-cost Pareto frontier through recursive harness self-improvement.
On Terminal-Bench~2, each round of evolution \textbf{\textit{improves the retained Pareto frontier}}, yielding higher pass rates at lower costs.
For example, the pass rate of the heavy mode rises from \textbf{74.6\%} in epoch~1 to a mean of \textbf{80.8\% }in epoch~3, while mean cost per task falls from \textbf{\$0.91} to \textbf{\$0.72}.
The lite and medium modes also maintain their pass rates from epoch~2 to~3, while their mean costs fall from \$0.61 to \$0.43 and from \$0.68 to \$0.56, respectively.
The benefits of additional evolution epochs are particularly clear on SWE-Bench Pro, where a single run may raise pass rate \textbf{\textit{at the expense of higher cost}}.
In the medium mode, pass rate rises from 70.0\% in epoch~1 to 77.0\% in epoch~2, but mean cost also rises from \$1.57 to \$1.66.
Only in epoch~3 does it improve both metrics, with pass rate rising to 79.0\% and mean cost falling to \$1.26 per task.
On ProgramBench, the fourth epoch likewise lowers costs across all three modes while maintaining similar pass rates.
These patterns highlight an advantage of \textbf{recursive} self-improvement over \textbf{single-round} evolution.

\subsection{Ablation Study}
\label{sec:ablation-study}

\begin{wraptable}{r}{0.60\textwidth}
\vspace{\dimexpr-\intextsep-1pt\relax}
\setlength{\abovecaptionskip}{13pt}
\centering
\caption{Ablations with GPT-5.5 and Codex.
In the RHO variants, \emph{+ dynamic routing} selects reasoning effort from task descriptions, \emph{+ budget hooks} append budget reminders to tool results and \emph{+ low reasoning effort} fixes effort at low.}
\label{tab:ablation-study}
\vspace{4pt}
\small
\definecolor{ablationTableHeader}{rgb}{0.94588,0.95764,0.96883}
\definecolor{ablationTableOurs}{rgb}{0.92587,0.97171,0.99509}
\setlength{\dashlinedash}{2pt}
\setlength{\dashlinegap}{1.5pt}
\renewcommand{\arraystretch}{1.12}
\setlength{\tabcolsep}{2.5pt}
\resizebox{\linewidth}{!}{%
\begin{tabular}{l*{6}{>{\centering\arraybackslash}p{3.1em}}}
\toprule
\rowcolor{ablationTableHeader}
& \multicolumn{3}{c}{\textbf{Terminal-Bench 2}}
& \multicolumn{3}{c}{\textbf{SWE-Bench Pro}} \\
\rowcolor{ablationTableHeader}
\multirow{-2}{*}{\textbf{Method}}
& \makebox[\linewidth][c]{\textbf{P(\%)}$\;\uparrow$} & \makebox[\linewidth][c]{\textbf{C(\$)}$\;\downarrow$} & \makebox[\linewidth][c]{\textbf{CoP(\$)}$\;\downarrow$}
& \makebox[\linewidth][c]{\textbf{P(\%)}$\;\uparrow$} & \makebox[\linewidth][c]{\textbf{C(\$)}$\;\downarrow$} & \makebox[\linewidth][c]{\textbf{CoP(\$)}$\;\downarrow$} \\
\midrule
RHO & 76.3 & 1.09 & 1.43 & 80.0 & 2.68 & 3.35 \\
RHO + dynamic routing & 76.3 & 0.92 & 1.21 & 76.0 & 2.13 & 2.80 \\
RHO + budget hooks & 76.3 & 0.92 & 1.21 & 75.0 & 1.75 & 2.33 \\
RHO + low reasoning effort & 62.7 & 0.45 & 0.72 & 59.0 & 1.08 & 1.83 \\
\hdashline
\rowcolor{ablationTableOurs}
\textbf{HarnessPro-Lite} & 72.9 & 0.43 & 0.59 & 72.3 & 0.80 & \textbf{1.11} \\
\quad w/o planner & 72.9 & \textbf{0.41} & \textbf{0.56} & 70.0 & 0.98 & 1.40 \\
\quad w/o test synthesizer & 67.8 & 0.67 & 0.99 & 64.0 & \textbf{0.78} & 1.22 \\
\hdashline
\rowcolor{ablationTableOurs}
\textbf{HarnessPro-Medium} & 76.3 & 0.56 & 0.73 & 79.0 & 1.26 & 1.59 \\
\quad w/o planner & 74.6 & 0.61 & 0.82 & 73.0 & 1.42 & 1.95 \\
\quad w/o test synthesizer & 72.9 & 0.82 & 1.12 & 69.0 & 1.37 & 1.99 \\
\hdashline
\rowcolor{ablationTableOurs}
\textbf{HarnessPro-Heavy} & \textbf{80.8} & 0.72 & 0.89 & \textbf{80.7} & 1.97 & 2.44 \\
\quad w/o planner & 79.7 & 1.04 & 1.30 & 79.0 & 2.12 & 2.68 \\
\quad w/o test synthesizer & 74.6 & 0.88 & 1.18 & 72.0 & 1.62 & 2.25 \\
\bottomrule
\end{tabular}%
}
\vspace{-4pt}
\end{wraptable}

Table~\ref{tab:ablation-study} examines the contributions of main components of \method{}.
We also report Cost-of-Pass (CoP)~\citep{erol2026costofpass}, the total execution cost divided by the number of successful tasks.
Removing the planner reduces performance and \textbf{increases cost} in most experiments.
For example, on SWE-Bench Pro, the pass rate of the medium mode drops from 79.0\% to 73.0\%, while mean cost rises from \$1.26 to \$1.42 per task.
Removing the test synthesizer has a \textbf{larger impact}, with pass rates dropping by \textbf{3.4--10.0} percentage points across both benchmarks and all three modes.
These performance drops highlight the value of our multi-agent framework.
Beyond these component ablations, we also examine whether RHO can achieve similar gains by simply adopting individual strategies from our evolved harnesses.
As shown in Table~\ref{tab:ablation-study}, adding either dynamic routing or budget hooks to RHO reduces cost, but can also \textbf{degrade} task performance.
On SWE-Bench Pro, both variants reduce cost but achieve lower pass rates than RHO's 80.0\%.
Uniformly using low reasoning effort yields larger cost savings but also substantially degrades performance.
Overall, all three RHO variants are Pareto-dominated by harnesses evolved by \method{}, which achieve higher pass rates at lower cost.

\section{Conclusion}
\label{sec:conclusion}

We introduced \method{}, a label-free approach to recursive harness self-improvement that improves task performance while reducing execution cost.
To this end, it coordinates trajectory diagnosis and harness modification through a multi-agent framework, with a dedicated planner proposing update plans and LLM-generated tests guiding candidate evaluation.
Rather than producing a single harness, \method{} offers \textbf{cost-focused} (lite), \textbf{balanced} (medium), and \textbf{performance-focused} (heavy) alternatives to meet different deployment budgets.
Experiments across three benchmarks show that \method{} can \textbf{\textit{improve task performance}} while \textbf{\textit{reducing execution cost}}, with additional benefits in \textbf{latency}.
Further analysis attributes these gains to the \textbf{synergy} of multiple evolved mechanisms and shows sustained progress through recursive self-improvement.
These findings show that performance and efficiency can improve together, making cost-aware evolution a promising direction for practical recursive self-improvement in harness optimization.

\subsubsection*{AI Use Statement}
Generative AI tools were used to assist with software implementation and the drafting and editing of this manuscript. Separately, LLMs are integral to the method studied in this work: they analyze execution trajectories, propose and implement harness updates, and synthesize tests during harness evolution. All AI-assisted code and text were manually reviewed, and all reported results were checked by the authors.

\subsubsection*{Reproducibility Statement}
The full evolution procedure and optimization objective are described in Sections~\ref{sec:problem-setting} and~\ref{sec:method}, with pseudocode in Algorithm~\ref{alg:harnesspro}. Section~\ref{sec:experimental-setup} specifies the benchmarks, model and agent versions, rollout counts, evolution epochs, selection hyperparameters, and evaluation metrics. The appendix provides the evolution instructions, token-pricing assumptions, and representative artifacts from the frozen evolved harnesses.

\bibliography{references}

@inproceedings{yehudai2026survey,
  title     = {A Survey on Evaluation of {LLM}-based Agents},
  author    = {Yehudai, Asaf and Eden, Lilach and Li, Alan and Uziel, Guy and Zhao, Yilun and Bar-Haim, Roy and Cohan, Arman and Shmueli-Scheuer, Michal},
  booktitle = {Findings of the Association for Computational Linguistics: ACL 2026},
  pages     = {26690--26714},
  year      = {2026},
  doi       = {10.18653/v1/2026.findings-acl.1330},
  url       = {https://aclanthology.org/2026.findings-acl.1330/}
}

@misc{openai2026agentarchitecture,
  author       = {{OpenAI}},
  title        = {Agents {API} Architecture},
  year         = {2026},
  howpublished = {\url{https://developers.openai.com/api/docs/guides/agents-api/architecture/}},
  note         = {Accessed: 2026-09-12}
}

@misc{grace2026demystifying,
  title        = {Demystifying evals for {AI} agents},
  author       = {Grace, Mikaela and Hadfield, Jeremy and Olivares, Rodrigo and De Jonghe, Jiri},
  howpublished = {Anthropic Engineering},
  year         = {2026},
  month        = jan,
  note         = {Published January 9, 2026},
  url          = {https://www.anthropic.com/engineering/demystifying-evals-for-ai-agents}
}

@inproceedings{zhang2026dgm,
  title     = {Darwin {Godel} Machine: Open-Ended Evolution of Self-Improving Agents},
  author    = {Zhang, Jenny and Hu, Shengran and Lu, Cong and Lange, Robert and Clune, Jeff},
  booktitle = {International Conference on Learning Representations},
  year      = {2026},
  url       = {https://openreview.net/forum?id=43n76kz3F5}
}

@inproceedings{zhang2025aflow,
  title     = {{AFlow}: Automating Agentic Workflow Generation},
  author    = {Zhang, Jiayi and Xiang, Jinyu and Yu, Zhaoyang and Teng, Fengwei and Chen, Xionghui and Chen, Jiaqi and Zhuge, Mingchen and Cheng, Xin and Hong, Sirui and Wang, Jinlin and Zheng, Bingnan and Liu, Bang and Luo, Yuyu and Wu, Chenglin},
  booktitle = {International Conference on Learning Representations},
  year      = {2025},
  url       = {https://arxiv.org/abs/2410.10762}
}

@article{lin2026ahe,
  title   = {Agentic Harness Engineering: Observability-Driven Automatic Evolution of Coding-Agent Harnesses},
  author  = {Lin, Jiahang and Liu, Shichun and Pan, Chengjun and Lin, Lizhi and Dou, Shihan and Xi, Zhiheng and Huang, Xuanjing and Yan, Hang and Han, Zhenhua and Gui, Tao and Jiang, Yu-Gang},
  journal = {arXiv preprint arXiv:2604.25850},
  year    = {2026},
  url     = {https://arxiv.org/abs/2604.25850}
}

@article{chen2026harnessfix,
  title   = {From Failed Trajectories to Reliable {LLM} Agents: Diagnosing and Repairing Harness Flaws},
  author  = {Chen, Mengzhuo and Wang, Junjie and Liu, Zhe and Wang, Yawen and Zheng, Haiming and Wang, Qing},
  journal = {arXiv preprint arXiv:2606.06324},
  year    = {2026},
  url     = {https://arxiv.org/abs/2606.06324}
}

@article{chen2026harnessx,
  title   = {{HarnessX}: A Composable, Adaptive, and Evolvable Agent Harness Foundry},
  author  = {Chen, Tingyang and Lu, Shuo and Zhao, Kang and Meng, Weicheng and Teng, Hanlin and Li, Tianhao and Li, Chao and Liu, Xule and Liang, Jian and Zhang, Zhizhong and Xie, Yuan and Qu, Heng and Shao, Kun and Luan, Jian},
  journal = {arXiv preprint arXiv:2606.14249},
  year    = {2026},
  url     = {https://arxiv.org/abs/2606.14249}
}

@article{pan2026rho,
  title   = {Evolving Agents in the Dark: Retrospective Harness Optimization via Self-Preference},
  author  = {Pan, Wenbo and Liu, Shujie and Lin, Chin-Yew and Zeng, Jingying and Tang, Xianfeng and Zhou, Xiangyang and Lu, Yan and Jia, Xiaohua},
  journal = {arXiv preprint arXiv:2606.05922},
  year    = {2026},
  url     = {https://arxiv.org/abs/2606.05922}
}

@article{marler2004survey,
  title   = {Survey of multi-objective optimization methods for engineering},
  author  = {Marler, R. T. and Arora, J. S.},
  journal = {Structural and Multidisciplinary Optimization},
  volume  = {26},
  number  = {6},
  pages   = {369--395},
  year    = {2004},
  doi     = {10.1007/s00158-003-0368-6}
}

@inproceedings{pan2026contextpilot,
  title     = {{ContextPilot}: Teaching Agents for Proactive Context Management via Fine-grained {RL}},
  author    = {Pan, Zhuoshi and Pei, Qizhi and Lu, Junru and Lin, Honglin and Zhao, H. Vicky and Yin, Di and Sun, Xing},
  booktitle = {Proceedings of the 2026 Conference on Empirical Methods in Natural Language Processing},
  month     = nov,
  year      = {2026},
  address   = {Budapest, Hungary},
  publisher = {Association for Computational Linguistics}
}

@article{yang2026skillopt,
  title   = {{SkillOpt}: Executive Strategy for Self-Evolving Agent Skills},
  author  = {Yang, Yifan and Gong, Ziyang and Huang, Weiquan and Yang, Qihao and Zhou, Ziwei and Huang, Zisu and Li, Yan and Gao, Xuemei and Dai, Qi and Liu, Bei and Qiu, Kai and Yang, Yuqing and Chen, Dongdong and Yang, Xue and Luo, Chong},
  journal = {arXiv preprint arXiv:2605.23904},
  year    = {2026},
  url     = {https://arxiv.org/abs/2605.23904}
}

@article{gong2026skillmoo,
  title   = {{SkillMOO}: Multi-Objective Optimization of Agent Skills for Software Engineering},
  author  = {Gong, Jingzhi and Gu, Ruizhen and Fei, Zhiwei and Cao, Yazhuo and Twist, Lukas and Geiger, Alina and Han, Shuo and Sobania, Dominik and Sarro, Federica and Zhang, Jie M.},
  journal = {arXiv preprint arXiv:2604.09297},
  year    = {2026},
  url     = {https://arxiv.org/abs/2604.09297}
}

@article{wang2026frames,
  title   = {{FRAMES}: Guarded and Dual-Objective Skill Evolution for Agents in Policy-Governed Enterprise Workflows},
  author  = {Wang, Xuhui and Shu, Ruoqi and Dan, Chen and Xu, Tianhua and Luo, Mengxi and Mai, Yanming and Wan, Bo},
  journal = {arXiv preprint arXiv:2608.01772},
  year    = {2026},
  url     = {https://arxiv.org/abs/2608.01772}
}

@article{lei2026memorycpt,
  title   = {{MemoryCPT}: An End-to-End Agent Memory Framework for Cost-Performance Trade-off},
  author  = {Lei, Songxin and Ouyang, Kun and Ruan, Weilin and Wu, Yuqian and Guo, Zhijiang and Sun, Yushi and Tsung, Fugee},
  journal = {arXiv preprint arXiv:2608.04843},
  year    = {2026},
  url     = {https://arxiv.org/abs/2608.04843}
}

@article{liu2025bats,
  title   = {Budget-Aware Tool Use Enables Effective Agent Scaling},
  author  = {Liu, Tengxiao and Wang, Zifeng and Miao, Jin and Hsu, I-Hung and Yan, Jun and Chen, Jiefeng and Han, Rujun and Xu, Fangyuan and Chen, Yanfei and Jiang, Ke and Daruki, Samira and Liang, Yi and Wang, William Yang and Pfister, Tomas and Lee, Chen-Yu},
  journal = {arXiv preprint arXiv:2511.17006},
  year    = {2025},
  url     = {https://arxiv.org/abs/2511.17006}
}

@article{gong2026compas,
  title   = {{COMPAS}: Difficulty-Aware Joint Search for Optimizing Code Generation},
  author  = {Gong, Jingzhi and Zhang, Jie M. and Jahangirova, Gunel and Huang, Dong and Mousavi, Mohammad Reza and Harman, Mark},
  journal = {arXiv preprint arXiv:2608.04336},
  year    = {2026},
  url     = {https://arxiv.org/abs/2608.04336}
}

@article{kulesza2012dpp,
  title   = {Determinantal Point Processes for Machine Learning},
  author  = {Kulesza, Alex and Taskar, Ben},
  journal = {Foundations and Trends in Machine Learning},
  volume  = {5},
  number  = {2--3},
  pages   = {123--286},
  year    = {2012}
}

@misc{tbench2025,
  title  = {{Terminal-Bench}: A Benchmark for {AI} Agents in Terminal Environments},
  author = {{The Terminal-Bench Team}},
  year   = {2025},
  month  = apr,
  url    = {https://github.com/laude-institute/terminal-bench}
}

@inproceedings{deng2026swebenchpro,
  title     = {{SWE-Bench Pro}: Can {AI} Agents Solve Long-Horizon Software Engineering Tasks?},
  author    = {Deng, Xiang and Da, Jeff and Pan, Edwin and He, Yannis Yiming and Ide, Charles and Garg, Kanak and Lauffer, Niklas and Park, Andrew and Rane, Chetan and Sampath, Karmini and Krishnan, Maya and Kundurthy, Srivatsa and Hendryx, Sean and Wang, Zifan and Zhang, Chen Bo Calvin and Jacobson, Noah and Liu, Bing and Kenstler, Brad},
  booktitle = {International Conference on Machine Learning},
  year      = {2026},
  url       = {https://icml.cc/virtual/2026/poster/61047}
}

@article{yang2026programbench,
  title   = {{ProgramBench}: Can Language Models Rebuild Programs From Scratch?},
  author  = {Yang, John and Lieret, Kilian and Ma, Jeffrey and Thakkar, Parth and Pedchenko, Dmitrii and Sootla, Sten and McMilin, Emily and Yin, Pengcheng and Hou, Rui and Synnaeve, Gabriel and Yang, Diyi and Press, Ofir},
  journal = {arXiv preprint arXiv:2605.03546},
  year    = {2026},
  url     = {https://arxiv.org/abs/2605.03546}
}

@inproceedings{suzgun2026dynamiccheatsheet,
  title     = {Dynamic Cheatsheet: Test-Time Learning with Adaptive Memory},
  author    = {Suzgun, Mirac and Yuksekgonul, Mert and Bianchi, Federico and Jurafsky, Dan and Zou, James},
  booktitle = {Proceedings of the 19th Conference of the European Chapter of the Association for Computational Linguistics (Volume 1: Long Papers)},
  publisher = {Association for Computational Linguistics},
  pages     = {7080--7106},
  year      = {2026},
  doi       = {10.18653/v1/2026.eacl-long.333},
  url       = {https://aclanthology.org/2026.eacl-long.333/}
}

@inproceedings{ouyang2026reasoningbank,
  title     = {{ReasoningBank}: Scaling Agent Self-Evolving with Reasoning Memory},
  author    = {Ouyang, Siru and Yan, Jun and Hsu, I-Hung and Chen, Yanfei and Jiang, Ke and Wang, Zifeng and Han, Rujun and Le, Long and Daruki, Samira and Tang, Xiangru and Tirumalashetty, Vishy and Lee, George and Rofouei, Mahsan and Lin, Hangfei and Han, Jiawei and Lee, Chen-Yu and Pfister, Tomas},
  booktitle = {The Fourteenth International Conference on Learning Representations},
  year      = {2026},
  url       = {https://openreview.net/forum?id=jL7fwchScm}
}

@article{lin2025sleeptime,
  title   = {Sleep-time Compute: Beyond Inference Scaling at Test-time},
  author  = {Lin, Kevin and Snell, Charlie and Wang, Yu and Packer, Charles and Wooders, Sarah and Stoica, Ion and Gonzalez, Joseph E.},
  journal = {arXiv preprint arXiv:2504.13171},
  year    = {2025},
  url     = {https://arxiv.org/abs/2504.13171}
}

@inproceedings{erol2026costofpass,
  title     = {Cost-of-Pass: An Economic Framework for Evaluating Language Models},
  author    = {Erol, Mehmet Hamza and El, Batu and Suzgun, Mirac and Yuksekgonul, Mert and Zou, James},
  booktitle = {The Fourteenth International Conference on Learning Representations},
  year      = {2026},
  url       = {https://openreview.net/forum?id=vC9S20zsgN}
}

@misc{szymkowiak2026rtk,
  title        = {{RTK}: {Rust Token Killer}},
  author       = {Szymkowiak, Patrick},
  year         = {2026},
  howpublished = {GitHub repository},
  url          = {https://github.com/rtk-ai/rtk},
  note         = {Accessed September 20, 2026}
}
\bibliographystyle{iclr2027_conference}

\clearpage
\appendix
\ifdefined\nolinenumbers\nolinenumbers\fi
\section{Implementation Details}
\label{apx:implement-details}

\subsection{DPP-Based Coreset Selection}
\label{apx:dpp-coreset}

We select the training coreset once, before harness evolution. Following RHO~\citep{pan2026rho}, let $\mathcal U=\{1,\ldots,M\}$ index the candidate tasks. For each task $x_i$, we collect a trajectory under the initial harness. A language-model judge reads the task description and trajectory, then assigns a difficulty score $r_i\in[0,10]$ and an abstract failure-mode fingerprint $\phi_i$. The fingerprint omits task-specific names so that similarity reflects shared challenges rather than superficial project details.

Let $u_i$ be the unit-normalized embedding of $\phi_i$. We compute cosine similarity and construct a difficulty-weighted DPP kernel:
\begin{equation}
\begin{aligned}
S_{ij} &= u_i^\top u_j, \qquad q_i=\max(r_i/10,\varepsilon),\\
w_i &= \left(q_i/\max_j q_j\right)^\alpha,
\qquad \alpha=\frac{\theta}{2(1-\theta)},\\
K &= \operatorname{diag}(w)\,S\,\operatorname{diag}(w).
\end{aligned}
\label{eq:dpp-coreset-kernel}
\end{equation}
We use $\varepsilon=0.1$ and $\theta=0.7$, as in RHO. The difficulty weights favor challenging tasks, while the determinant of a subset's kernel penalizes redundant fingerprints. For a fixed coreset size $k$, the corresponding $k$-DPP assigns each $k$-element subset $Y\subseteq\mathcal U$ probability~\citep{kulesza2012dpp}
\begin{equation}
\Pr_k(Y)
=\frac{\det(K_Y)}{\sum_{Z\subseteq\mathcal U,\,|Z|=k}\det(K_Z)},
\label{eq:dpp-coreset-probability}
\end{equation}
where $K_Y$ is the principal submatrix indexed by $Y$. We do not sample from this distribution. Instead, we approximate its most probable subset by greedily adding the task that maximizes the resulting determinant:
\begin{equation}
\begin{aligned}
i_s &= \arg\max_{i\in\mathcal U\setminus Y_s}
\det(K_{Y_s\cup\{i\}}),\\
Y_{s+1} &= Y_s\cup\{i_s\},\qquad Y_0=\emptyset.
\end{aligned}
\label{eq:dpp-coreset-greedy}
\end{equation}
We stop after $k=10$ tasks and keep this coreset fixed across all evolution epochs. Neither official test results nor validation outcomes enter coreset selection.

\subsection{Additional Implementation Details}

\paragraph{Reasoning effort and mode weights.}
The default reasoning effort for GPT-5.5 is high on all three benchmarks. For GLM-5.2, the default is max on Terminal-Bench~2 and SWE-Bench Pro, but high on ProgramBench, where max tends to induce extensive reasoning and tool use, substantially lengthening execution.
For the selection rule in Equation~\ref{eq:mode-selection}, the performance-cost weights $(\alpha_m,\beta_m)$ are $(1,2)$ for lite, $(1,1)$ for medium, and $(2,1)$ for heavy. These weights remain fixed across epochs.

\paragraph{Generation settings.}
Our Codex configurations use the Responses API without explicitly setting temperature, top\_p, or max\_output\_tokens. When used through Codex, GPT-5.5 therefore uses its default sampling settings. The model supports up to 128,000 output tokens per response.\footnote{\url{https://developers.openai.com/api/docs/models/gpt-5.5}}
For GLM-5.2, Terminal-Bench~2 and SWE-Bench Pro use temperature 1.0, top-p 0.95, and a per-request \texttt{max\_tokens} of 131,072. On ProgramBench, GLM-5.2 runtime uses temperature 1.0, top-p 1.0, and \texttt{max\_tokens} of 32,000, with a 2,000-turn limit.

\paragraph{Solver and grading budgets.}
On Terminal-Bench~2, GPT-5.5 uses each task's native solver limit (600--3,600 seconds on the test set). On SWE-Bench Pro, the solver time limit is 1200 seconds for GPT-5.5 and 7,200 for GLM-5.2, and the grader limit is 3,600 seconds for both models. On ProgramBench, the solver budget is 7,200 seconds, separate from the 14,400 second outer limit for each grading branch. These wall-clock limits are distinct from harness-level tool-call budgets.

\paragraph{Task environments.}
Terminal-Bench~2 and ProgramBench allocate 8 CPUs and 32~GiB of memory per solver. SWE-Bench Pro runs solvers in host workspaces without an explicit per-task CPU or memory limit. Candidate evaluation uses one rollout per validation task, and final testing is performed only after evolution. Infrastructure failures are excluded and retried to avoid inaccurate results.

\paragraph{Token pricing.}
Table~\ref{tab:token-pricing} reports the public API prices used to convert token usage into execution cost. We use the short-context price tier for GPT-5.5 because all requests remain below 272K context tokens. Prices follow the official OpenAI\footnote{\url{https://developers.openai.com/api/docs/pricing}} and Z.ai\footnote{\url{https://docs.z.ai/guides/overview/pricing}} pricing pages.

\begin{table}[!t]
  \centering
  \caption{Token prices used to compute execution costs.}
  \label{tab:token-pricing}
  \small
  \begin{tabular}{@{}lccc@{}}
    \toprule
    \textbf{Model} & \textbf{Uncached input} & \textbf{Cached input} & \textbf{Output} \\
    \midrule
    GPT-5.5 ($<272$K context) & 5.00~\$/M & 0.50~\$/M & 30.00~\$/M \\
    GLM-5.2 & 1.40~\$/M & 0.26~\$/M & 4.40~\$/M \\
    \bottomrule
  \end{tabular}
\end{table}

\clearpage
\section{Evolution Instructions}
\label{app:evolution-prompt}

\begin{artifacttextbox}[lst:evolution-instruction-lite]{Cost-focused evolution instruction ($I_1$)}
Prioritize reducing average execution cost while preserving task correctness and
reliable completion. Seek genuine efficiency gains, not lower bills obtained by
skipping required work, abandoning difficult tasks, or weakening validation.

Use the diagnostics, trajectories, and candidate implementations to distinguish
necessary work from avoidable expense. Look for recurring sources of redundant
exploration, repeated context, or low-value verification. Consider the total cost of
execution, including unsuccessful attempts and any overhead introduced by the proposed
change; fewer steps or shorter outputs alone do not establish an improvement.

Possible directions include adapting reasoning effort to individual public tasks,
managing context and tool feedback economically, and improving convergence or recovery.
Preserve the information and checks needed for dependable results. These are optional
search directions, not a checklist: derive any mechanism from the available evidence,
without assuming a routing rubric, fixed thresholds, profile taxonomy, or per-task rules.

Propose a focused, generalizable plan rather than accumulating optimizations. Explain
which observed expense each change addresses, why correctness should be preserved,
and what regression risks remain. Distinguish expected savings from measured results;
the fixed external evaluator, not the plan's claims, determines whether quality is
maintained and cost is reduced.
\end{artifacttextbox}

\begin{artifacttextbox}[lst:evolution-instruction-medium]{Balanced evolution instruction ($I_2$)}
Seek a balanced improvement in task reliability and average execution cost. Treat
correctness and efficiency as joint objectives: do not pursue savings by neglecting
required behavior, or assume that additional computation necessarily improves quality.
A balanced plan should improve the trade-off, not simply average the other directions.

Use the diagnostics, trajectories, and candidate implementations to identify where
effort contributes to successful completion and where it is redundant. Compare peer
candidates to understand their strengths, weaknesses, and interactions, rather than
copying their mechanisms wholesale. Prefer changes that remove low-value work while
preserving or strengthening the reasoning and validation that matter.

Possible directions include task-adaptive reasoning effort, concise but recoverable
context, effective use of tool feedback, and verification or recovery guided by observed
uncertainty. These are optional hypotheses, not prescribed designs. Infer an appropriate
approach from the evidence; no routing rubric, fixed thresholds, profile names, or
per-task rules are supplied. Account for the overhead and complexity of combining changes.

Propose a coherent, generalizable plan and explain its expected quality and cost effects
together. Make uncertain trade-offs and regression risks explicit, including why any
added work is worthwhile. Keep predicted benefits separate from established findings
and let the fixed external evaluator determine whether the joint improvement is real.
\end{artifacttextbox}

\clearpage
\begin{artifacttextbox}[lst:evolution-instruction-heavy]{Performance-focused evolution instruction ($I_3$)}
Prioritize improving task success while continuing to eliminate avoidable expense.
Focus on reliable completion and correct behavior, not more elaborate execution for
its own sake. Performance-focused evolution should invest effort where it can address
meaningful failures, rather than increase computation uniformly across tasks.

Use the diagnostics, trajectories, and candidate implementations to identify
consequential failure patterns and the limitations behind them. Examine both successful
and unsuccessful executions to distinguish useful reasoning or validation from activity
that merely adds cost. Draw on peer candidates when their behavior provides relevant
evidence, but do not assume that a more complex harness is a more capable one.

Possible directions include adapting reasoning effort to the public task, selective
deeper investigation or independent verification, and stronger context or recovery
management. These are optional hypotheses, not an implementation recipe. Derive the
actual design from evidence; no routing rubric, fixed thresholds, mandatory review
schedule, or per-task rules are supplied. Preserve efficient behavior where additional
work has no demonstrated purpose.

Propose a focused, generalizable plan that explains which failures it aims to prevent
and why the changes should help. If additional work is necessary, state the expected
quality benefit, cost implications, and remaining risks without presenting predictions
as measured gains. The fixed external evaluator determines whether the improvement
justifies the resulting quality-cost trade-off.
\end{artifacttextbox}
\clearpage

\clearpage
\section{Representative Multi-Agent Artifacts}
\subsection{Trajectory Diagnoses}
\label{app:diagnosis-example}

These lightly edited analyzer outputs retain issue-level findings from three rollouts of each Terminal-Bench~2 training task. Outcomes use fixed LLM-generated tests. 

\paragraph{Caffe--CIFAR-10.}
All three runs failed before training. The analyzer separates a shared build bottleneck from run-specific stalls and a potential downstream risk.

\begin{diagnosisbox}[case:diagnosis-caffe]{Caffe--CIFAR-10: incomplete training}
Task: Build CPU-only Caffe and train CIFAR-10 for 500 iterations.
Outcome: No run completed conversion or training, saved a model or log,
  or verified accuracy.

Issue 1: All three workflows ended before training.
  Problem: Run 0 was fixing a protobuf build issue; run 1 had built
    Caffe but was downloading CIFAR-10; run 2 was still rebuilding after
    an out-of-memory (OOM) failure.
  Cost impact: Long installs, builds, and downloads produced no final
    training artifacts.
  Evidence: run-0-trace-0.jsonl:89-97; run-1-trace-0.jsonl:96-111;
    run-2-trace-0.jsonl:74-80.

Issue 2: High build parallelism caused OOM failures in every run.
  Problem: Runs 0 and 1 used make all -j$(nproc), and run 2 used
    cmake --build build --parallel 7. OOM-killed compiler processes
    prompted lower-parallelism retries.
  Quality impact: Runs 0 and 2 still lacked a successful build at the
    trace end, delaying data preparation and training.
  Cost impact: Failed compile passes and rebuilds wasted time.
  Evidence: run-0-trace-0.jsonl:71-89; run-1-trace-0.jsonl:90-96;
    run-2-trace-0.jsonl:61-80.

Issue 3: The most advanced run stalled on the CIFAR-10 download.
  Problem: Run 1 interrupted a slow single-stream download, installed
    aria2, and ended while its replacement parallel download was ongoing.
  Quality impact: Conversion, mean generation, training, and model
    verification were not reached despite a completed Caffe build.
  Cost impact: The interrupted download, package install, and retry
    added post-build time without completing data preparation.
  Evidence: run-1-trace-0.jsonl:96-111.

Issue 4: A likely downstream data-preparation blocker (not observed).
  Problem: Run 0 configured Caffe without OpenCV, although the stock
    CIFAR pipeline calls compute_image_mean, which requires USE_OPENCV.
  Quality impact: Had run 0 reached this step, mean generation would
    likely have failed; the trace does not show such a failure.
  Cost impact: An OpenCV-enabled rebuild or alternative preparation path
    would likely have been needed after the existing build work.
  Evidence: run-0-trace-0.jsonl:60-69; run-1-trace-0.jsonl:69;
    run-2-trace-0.jsonl:36-37.

Issue 5: Run 0 recovered from an incorrect source tag.
  It tried the nonexistent 1.0.0 branch, inspected tags, then used 1.0;
  this added commands before the build but no lasting quality loss.
  Evidence: run-0-trace-0.jsonl:27-42.
\end{diagnosisbox}

\clearpage
\paragraph{Mailman.}
In contrast, all three Postfix/Mailman~3 configurations passed the generated tests. The diagnosis separates useful end-to-end verification from broad exploration, redundant checks, and two run-specific detours that increased cost without changing the final outcome.

\begin{diagnosisbox}[case:diagnosis-mailman]{Mailman: successful but costly execution}
Task: Configure a mailing list with join, leave, and post delivery.
Outcome: All three rollouts passed the generated tests.

Issue 1: Exploration went well beyond the relevant configuration.
  Problem: All runs dumped large Mailman/Postfix configs and package
    source/tests after the task and eval narrowed the relevant surface
    to mailman.cfg, Postfix local delivery, and Mailman LMTP maps.
  Quality impact: Some inspection helped, but the broader search added
    little evidence for final correctness.
  Cost impact: Trace totals reached 1,058,946, 1,763,487, and
    2,517,861 input tokens in runs 0, 1, and 2, respectively; broad
    dumps and repeated exploration were major contributors.
  Evidence: run-0-trace-0.jsonl:36-69,86-93,180;
    run-1-trace-0.jsonl:22-27,175; run-2-trace-0.jsonl:18-19,180.

Issue 2: Final evaluations repeated already covered behavior.
  Problem: Manual join/post/leave smoke tests usefully checked delivery
    and leave confirmation, but /app/eval.py was rerun after these and
    earlier evaluations had passed.
  Quality impact: Extra evals added little end-to-end coverage.
  Cost impact: Each eval created Unix users, exercised mail queues,
    and included retry sleeps.
  Evidence: run-0-trace-0.jsonl:149-153;
    run-1-trace-0.jsonl:158-173;
    run-2-trace-0.jsonl:102-122,167-170.

Issue 3: Run 2's smoke test temporarily broke mail delivery.
  Problem: Opening /var/mail/<user> with mailbox.mbox() before Postfix
    created it produced a root-owned mailbox, so confirmation mail bounced.
  Quality impact: The agent diagnosed this test-induced failure and
    reran the smoke test successfully; the final outcome was unaffected.
  Cost impact: Log/queue inspection, qfile decoding, reconnects, and
    another full smoke test followed.
  Evidence: run-2-trace-0.jsonl:123-152.

Issue 4: A service-control pipeline hung in run 1.
  Problem: A daemonized Mailman start inherited a pipe to sed, leaving
    the wrapper running after the services started.
  Quality impact: Services ran, so the final result was unaffected.
  Cost impact: The agent polled, inspected logs and processes, then
    terminated the wrapper before evaluation.
  Evidence: run-1-trace-0.jsonl:104-126.

Issue 5: A broad recipient-validation workaround carried a risk.
  Problem: Run 1 disabled SMTP-time local recipient validation after
    lookup failures; run 2 also ended with local_recipient_maps empty.
  Quality impact: Delivery passed for valid Unix users. Invalid local
    recipients might be accepted at SMTP time and fail later, a risk
    not observed in the tests.
  Cost impact: Two failed evals and map debugging preceded the fallback.
  Evidence: run-1-trace-0.jsonl:130-159;
    run-2-trace-0.jsonl:99-100,172-173.
\end{diagnosisbox}

\clearpage
\subsection{Evolution Plans}
\label{app:planner-examples}

The following Terminal-Bench~2 plans illustrate three performance-cost priorities. The medium and heavy plans start from the same initial harness and trajectory diagnoses. A later lite proposal for an edited descendant responds to a router that often chose high or xhigh. The listings retain the Planner's wording and structured fields, with line breaks added for readability. Some passages are omitted to save space.

\begin{diagnosisbox}[case:planner-balanced]{Medium plan: introduce task-adaptive routing (excerpt)}
objective: Improve the empty baseline harness by adding general solver
  guidance and lightweight execution support that preserves task-relevant
  verification while cutting repeated broad exploration, noisy command
  output, and overconfident local-success conclusions.
changes:
  README.md: Add concise solver guidance: start from the public task
    contract and required artifacts;
    (*@\textcolor{artifactExcerpt}{\ttfamily ... (content is excerpted to save space) ...}@*)
  runtime.json: Add a small runtime configuration referencing
    router_prompt.txt and setting tool_output_token_limit to a moderate
    value such as 12000, without adding a review session or replacing
    the model instructions.
  router_prompt.txt: Add a general cost-aware routing prompt that
    chooses the lowest reasoning_effort likely to complete the public
    task: low for bounded recovery or simple one-artifact tasks, medium
    for routine code/configuration with clear tests, high for multi-file
    implementations, builds, interpreters, protocol/integration work,
    numerical algorithms, or weak local oracles, and xhigh only for
    unusually open-ended tasks where failure would likely come from
    deep reasoning rather than environment setup. Require JSON with
    reasoning_effort and a short rationale.
  tools/guarded-run: Add an executable POSIX shell helper for long or
    noisy commands. It should run a command under a timeout, save full
    stdout/stderr to a log file, and print only exit status, elapsed
    time, log path, and bounded head/tail excerpts.
    (*@\textcolor{artifactExcerpt}{\ttfamily ... (content is excerpted to save space) ...}@*)
  tools/quick-probe: Add an executable POSIX shell helper that prints a
    compact one-time environment snapshot: cwd, top-level files, whether
    the directory is a git repo, and availability of common tools.
    (*@\textcolor{artifactExcerpt}{\ttfamily ... (content is excerpted to save space) ...}@*)
quality_impact: Established findings show failures and regressions came
  less from insufficient raw deliberation and more from shallow or
  misread validation, unfinished long-running workflows, environment
  assumptions, and task-integrity problems.
  (*@\textcolor{artifactExcerpt}{\ttfamily ... (content is excerpted to save space) ...}@*)
cost_impact:
  (*@\textcolor{artifactExcerpt}{\ttfamily ... (content is excerpted to save space) ...}@*)
  The router adds a small high-effort classification call, but should
  repay that overhead on simpler tasks by avoiding default high-effort
  solving while preserving higher effort for tasks where it is likely
  useful.
  (*@\textcolor{artifactExcerpt}{\ttfamily ... (content is excerpted to save space) ...}@*)
risks: The router may under-allocate effort for deceptively simple tasks,
  and the routing call can be net-negative on very small tasks. Output
  caps and guarded summaries can hide important diagnostics unless the
  solver follows the log-path recovery guidance.
  (*@\textcolor{artifactExcerpt}{\ttfamily ... (content is excerpted to save space) ...}@*)
\end{diagnosisbox}

\clearpage
\begin{diagnosisbox}[case:planner-performance]{Heavy plan: verify task artifacts (excerpt)}
objective: Add a small, general harness that steers solvers toward
  public-contract verification, bounded exploration, and cheaper command
  output while preserving flexibility for varied tasks.
  (*@\textcolor{artifactExcerpt}{\ttfamily ... (content is excerpted to save space) ...}@*)
changes:
  README.md: Create concise solver guidance: derive an acceptance
    checklist from the public task first; use only public task files and
    legitimate upstream documentation, never benchmark
    answers/graders/solutions; inspect files in narrow slices; check
    tool availability once; prefer Python fallback inspection in minimal
    containers; avoid dumping large logs, matrices, generated files, or
    whole source files; complete all running commands before final;
    verify required artifacts with semantic/end-to-end checks rather
    than only syntax, existence, or nonblank outputs.
  runtime.json: Create with a conservative tool_output_token_limit, for
    example 16000, to cap accidental giant outputs from package managers,
    build logs, tracebacks, git usage text, and file dumps while still
    leaving room for meaningful compiler or test failures.
  tools/run_logged.py: Add an executable Python helper with a shebang
    that runs a command with an optional timeout, writes the full
    stdout/stderr stream to a log file, and prints only command, exit
    code, elapsed time, and the final bounded tail. README should
    recommend it for apt/build/train/long-test commands so expensive
    workflows can preserve diagnostics without flooding context.
  tools/probe.py: Add an executable Python helper with a shebang for
    common low-dependency inspections: command-exists checks, limited
    head/tail display, hex/ascii byte previews, JSON validation, CSV
    column/row summaries, file size/hash, and simple BMP/PNG metadata
    where possible. This replaces repeated assumptions about
    unavailable utilities such as file, sqlite3, xxd, strings, and git.
quality_impact: Expected to improve reliability by making solvers
  explicitly map public requirements to final artifacts, require
  semantic verification before success claims, and avoid continuing
  from misleading checks such as command-not-found text hidden inside
  successful pipelines.
  (*@\textcolor{artifactExcerpt}{\ttfamily ... (content is excerpted to save space) ...}@*)
cost_impact: Expected to reduce avoidable token and latency costs from
  repeated broad file reads, package/build log dumps, missing-tool
  probes, noisy git failures, and redundant post-success inspection.
  (*@\textcolor{artifactExcerpt}{\ttfamily ... (content is excerpted to save space) ...}@*)
  There is no added second solver session or uniform extra reasoning
  pass.
risks: Guidance-only changes depend on solver compliance, and an output
  cap can hide useful detail if commands are not logged through the
  helper. The helper scripts must remain dependency-free and generic.
  More rigorous semantic verification can add some task-local execution
  cost, so the README should emphasize checks tied to public
  requirements rather than broad extra exploration.
\end{diagnosisbox}

\clearpage
\begin{diagnosisbox}[case:planner-lite]{Lite plan: make routing more cost-sensitive (excerpt)}
objective: Reduce solver and exploratory cost by correcting a router that
  was observed to choose high or xhigh for most sampled tasks, while
  retaining task-specific reasoning-effort selection and public-contract
  correctness.
changes:
  runtime.json: Keep router_prompt connected so each task still receives
    a dynamically selected reasoning_effort.
    (*@\textcolor{artifactExcerpt}{\ttfamily ... (content is excerpted to save space) ...}@*)
    Reduce tool_output_token_limit from 12000 to about 8000 to limit recurring
    transcript growth without removing useful diagnostics.
  router_prompt.txt: Rewrite the rubric to counter the observed
    high/xhigh bias: make medium the preferred starting point for bounded
    tasks with a plausible implementation path and practical checks,
    even when they mention several files, a build, data processing, or
    an unfamiliar domain. Choose low more readily when the requested
    change is concrete and its result can be checked directly. Escalate
    to high only for substantial unresolved reasoning or weak
    verification, and to xhigh only for exceptional cases where deeper
    reasoning is likely to change the solution.
    (*@\textcolor{artifactExcerpt}{\ttfamily ... (content is excerpted to save space) ...}@*)
  README.md: Revise guidance to reference helpers by absolute harness
    path, such as $HARNESSPRO_HARNESS/tools/guarded-run and
    $HARNESSPRO_HARNESS/tools/quick-probe, because relative tools/ paths
    failed from task working directories.
    (*@\textcolor{artifactExcerpt}{\ttfamily ... (content is excerpted to save space) ...}@*)
    stop extra stress testing
    once public-contract coverage plus one meaningful semantic check
    has passed.
  tools/guarded-run: Tighten defaults and output shaping: reduce the
    default excerpt size, add a per-line character cap while preserving
    the full log path, and keep timeout/status/elapsed reporting.
    (*@\textcolor{artifactExcerpt}{\ttfamily ... (content is excerpted to save space) ...}@*)
  tools/quick-probe: Extend the probe output to include
    HARNESSPRO_HARNESS, the absolute helper paths, and a compact
    availability table only.
    (*@\textcolor{artifactExcerpt}{\ttfamily ... (content is excerpted to save space) ...}@*)
quality_impact: Dynamic routing remains available for tasks that
  genuinely need high or xhigh reasoning, while the solver guidance
  strengthens public-contract and semantic validation. Favoring medium
  and, for directly checkable work, low should not sacrifice quality
  solely to save tokens; the router must still escalate when the
  approach or verification is genuinely uncertain.
cost_impact: The observed router usually chose high or xhigh despite
  adding its own high-reasoning turn, so it incurred fixed routing cost
  without much solver-effort reduction.
  (*@\textcolor{artifactExcerpt}{\ttfamily ... (content is excerpted to save space) ...}@*)
  Savings remain a hypothesis until measured on task rollouts.
risks: The observed high/xhigh choices may include genuinely difficult
  tasks; shifting toward medium or low could hurt quality if public
  descriptions hide hard reasoning or weak oracles.
  (*@\textcolor{artifactExcerpt}{\ttfamily ... (content is excerpted to save space) ...}@*)
  Keep high and xhigh available for substantial uncertainty, and
  compare per-task outcomes and costs before treating the revised
  rubric as an improvement.
\end{diagnosisbox}

\clearpage
\subsection{Test Synthesizer Configuration}
\label{app:test-synthesizer-setup}

Our test-synthesis agent uses an instruction and skill together with an MCP-based validation tool. The first two listings present selected passages from the deployed text, with output-schema fields condensed; the third summarizes the tool interface. Together, they show how the agent designs tests from public evidence, checks them against a frozen artifact, and reviews the suite without access to official tests or labels.

\begin{artifacttextbox}[lst:test-synth-instruction]{Test-synthesis instruction (excerpt)}
Use $empty-test-authoring. In this one Codex invocation, write ten distinct
executable cases, validate them on the frozen Empty product, inspect the results,
and self-review the final suite. No separate reviewer is part of this run.

## Evidence boundary
Read /work/public-task.json, relevant /public_inputs, and the read-only
/empty_product. The task description and public files define the contract;
Empty is only an implementation and execution-diagnostic context, never an
answer key. Public sample programs and evaluation helpers may be used.
...
Do not seek or use the target's hidden official tests, grader, solution,
grades, or earlier target suites, locally or online. Do not modify Empty.

## Test design
Inventory the public requirements and the actual input and output interfaces
before testing Empty. Cover distinct documented behaviors, valid alternatives,
boundaries, and justified combinations; do not invent extra obligations to
fill the fixed count.
...
For each central requirement, invoke the real delivered product and assert a
meaningful relation between its output and independent public evidence.
A file existing, successful process, shape, nonempty value, or plausible range
alone does not establish the required behavior.
...
Choose the strongest justified oracle available: compatible public reference,
known answer, necessary content property, or discriminating metamorphic
relation. Never derive expected constants, hashes, tolerated values, or
execution paths from Empty.

## Execute and self-review
Write the complete suite, then call validate_generated_tests on it. A tool
result with status=ok, passed=false is a valid execution with possible product
failures; making Empty pass is not the objective. infra_error is inconclusive:
inspect the traceback and public evidence to distinguish test/setup defects
from product exceptions under valid invocation. Repair only demonstrated test
defects, preserving the public requirement and legitimate Empty failures. Do
not skip cases, catch all exceptions, or tune expectations toward Empty.
After every executable edit, rerun the tool; the final executable files must
match a completed execution without infrastructure error. Run tests serially.
Self-review every case, expected value, and shared helper after the final run.
...
## Output contract
Write UTF-8 files under /work/output, at most 1 MiB per file and 4 MiB total:
- test_generated.py: exactly test_case_0001 through test_case_0010 as
  unittest.TestCase methods; no skipped or unconditional padding cases.
- case_manifest.json: expected_count: 10; each case has a meaningful
  description and kind behavioral, format, or unavailable_oracle.
- coverage.json: public evidence, requirements mapped to case IDs, checks
  performed, what was not checked, and limitations.
- repair-log.json: demonstrated defects, public evidence, edits, and rerun
  outcomes; use [] if no repair was needed.
- self-review.json: per-case assessment, reason, public evidence, suggested
  repair, and any essential or supplemental coverage gaps.
\end{artifacttextbox}

\begin{artifacttextbox}[lst:test-synth-skill]{Empty-test-authoring skill (excerpt)}
Read /work/public-task.json and relevant /public_inputs first. Inspect
/empty_product only to understand the frozen candidate's interfaces or to
diagnose execution. Its files correspond to original paths such as /app, but
the snapshot is not an oracle. The available public dependencies and fixture
mounts are listed in validation_environment in the task JSON. Disjoint
examples are read-only under /examples.

Write the suite in /work/output. Invoke the frozen_empty MCP tool
validate_generated_tests after writing executable tests and after each edit;
the tool executes the current suite in a fresh Empty container and returns
only synthetic diagnostics. If MCP discovery fails, send a JSON-RPC tools/call
request to python3 /opt/harnesspro/empty-tools/tools/mcp_server.py on stdin.
Keep tests and validation calls serial. The model has no Docker socket or
official grader.

Interpret status=ok, passed=false as a completed execution, not a request
to make Empty pass. Treat infra_error as unresolved until its cause is
identified. Repair only test defects supported by public evidence, record
the change in repair-log.json, and rerun the changed suite. Preserve valid
product failures and report any unresolved oracle or setup problem. Review
the final test design in the same invocation; do not launch another reviewer.

Public reference tools, if needed, are available in the authoring image;
version manifests are under /opt/harnesspro/reference-tools. Check that a
reference is compatible with the public runtime before using exact outputs.
The main instruction defines the ten-case output and review schema.
\end{artifacttextbox}

\begin{artifacttextbox}[lst:test-synth-tool]{Validation tool interface and response (summary)}
Tool: frozen_empty.validate_generated_tests
MCP discovery: tools/list advertises validate_generated_tests
Arguments: {} (no task-specific parameters; extra properties rejected)
Input: the current test suite in /work/output
Request: a validate_generated_tests operation sent through /empty-tool-rpc
Execution: a fresh container with the frozen Empty product
Response: JSON diagnostics from the suite execution, not official tests,
  grader feedback, or labels
Completed execution: status == ok, whether passed is true or false
Infrastructure failure: status == infra_error; MCP isError is true
Timeout: 7,200 seconds; an absent response raises an error rather than
  inventing a validation result
\end{artifacttextbox}

\section{Evolved Harness Artifacts}
\label{app:evolved-harnesses}

This appendix presents the medium configurations of the evolved harnesses for
Terminal-Bench~2, SWE-Bench Pro, and ProgramBench. For Terminal-Bench~2, we also
show how the lite and heavy modes differ from the medium configuration.
Long lines wrap, JSON whitespace is normalized, and closing Markdown inline-code
backticks are typeset as apostrophes where needed. The bundled files
contain the evaluated settings.
The directory listings show the evolved harness files.


\subsection{Terminal-Bench 2}
\label{app:tb2-artifacts}

Listing~\ref{lst:tb2-layout} shows the runtime files of the medium-mode
harness. A separate GPT-5.5 call uses the rubric to select an execution profile
from the public task description. The evaluator reads the policy settings and
generates a Codex configuration that loads the instruction file and registers
the hook.

\begin{artifactbox}[lst:tb2-layout]{Terminal-Bench 2 medium-mode harness runtime files}
runtime/
    model_instructions.md
    profile_rubric.md
    policy.json
    hook.py
\end{artifactbox}

\subsubsection{Medium Mode}
\label{app:tb2-medium}
The medium mode uses GPT-5.5 throughout, with five execution profiles spanning
low, medium, and high reasoning effort. The listings below present its
instructions, routing rubric, runtime hooks, and policy settings.

\begin{artifacttextbox}[lst:tb2-instructions]{Medium: \texttt{runtime/model\_instructions.md}}
You are Codex, a non-interactive software agent. Work autonomously until the
user's task is complete. Follow the task's integrity, safety, path, and output
requirements exactly. Use the available tools to inspect, edit, execute, and
verify; batch independent checks and avoid repeating unchanged reads. Preserve
unrelated files and avoid destructive operations outside the requested scope.
Do not replace a requested implementation or service with a mock, proxy, or
look-alike unless the task explicitly permits that substitution.
Do not spawn other agents or ask the user to perform routine steps. Keep
progress messages brief. Before finishing, leave every requested artifact at
its exact path and run the strongest allowed task-shaped verification. Report
the outcome concisely, including any genuine remaining uncertainty.
\end{artifacttextbox}

\begin{artifacttextbox}[lst:tb2-session]{Medium: \texttt{policy.json}: \texttt{session\_context}}
Harness guidance is already loaded; do not inventory harness/. Extract exact artifacts, paths, behavior, integrity restrictions, public interfaces, and final state from task/prompt.md. Use its host-to-container pattern. For black-box work use only public interfaces and implement general behavior. When the task names an API, wrapper, package, revision, runtime, or service, exercise that named interface first and do not silently replace it with a lower-level dependency, mock, proxy, or look-alike behavior; inspect the public signature or implementation after an interface error. A source-only or single-artifact deliverable should not acquire undeclared runtime helpers. Start with the smallest discriminating probe and batch independent checks. Bound downloads, external interaction, and potentially blocking commands well below the task watchdog; reserve time for final repair and verification. Validate requested behavior, not only superficial readiness signals. Run mutating workflow checks on disposable fixtures when possible; if the user is meant to perform the same workflow later, do not leave test data or state that collides with its documented preconditions. Preserve the real requested implementation and durable service state while removing only temporary verification artifacts. For quantitative work, identify units, variables, invariants, and the comparison objective before optimizing; exercise every publicly stated correctness dimension using a direct reference or invariant when one can be constructed. Keep temporary verification artifacts separate from final deliverables. Before finishing, compare exact contents and state with every explicit requirement. Avoid narration and unchanged rereads; do not skip necessary validation.
\end{artifacttextbox}

\begin{artifacttextbox}[lst:tb2-rubric]{Medium: \texttt{runtime/profile\_rubric.md}}
Choose exactly one profile from observable task characteristics. Optimize
expected success per dollar. Never use historical outcomes remembered from prior
runs or evaluations.

Profiles:

- `compact': low reasoning; converge near 4 tool calls and stop near 8.
- `lean': low reasoning; converge near 10 calls and stop near 16.
- `standard': medium reasoning; converge near 26 calls and stop near 38.
- `deep': high reasoning; converge near 30 calls and stop near 48.
- `research': high reasoning with about 40 calls for source reconstruction.

Evidence rules

- Use only validation, interaction, dependencies, artifacts, and constraints
  explicitly described in the task text. Never assume hidden-test behavior,
  previous attempts, tool output, runtime results, or historical scores.
- An undescribed validator has unknown quality, not weak quality. Unknown decisive
  evidence defaults to `standard'; missing information alone cannot justify
  another profile.
- A strong oracle must be explicitly local and repeatable and provide localized
  failure, quantitative error, candidate ranking, deterministic simulation, or a
  directly inspectable target. Compilation, launch, an open port, one end-to-end
  request, or whole-task pass/fail is weak only when the text explicitly limits
  feedback to that undifferentiated signal.
- Count distinct evidence-resolvable hypotheses, not files, prose, commands, or
  retries. Extra iterations are valuable only when described probes can
  distinguish hypotheses and change the plan.
- Include the stated watchdog and expected probe cost in the value of
  iteration. A precise oracle is not useful enough for `deep' when only a few
  full evaluations can fit and high reasoning does not create more evidence.

Apply the following gates in order and stop at the first match.

Gate 1: `compact'

Choose `compact' only when low marginal value is explicit and no faithful
decomposable oracle is described or structurally possible:

1. Long serial graphical, remote, virtualized, or manual interaction whose
   correctness depends on visual or hidden state.
2. Unavailable accounts, devices, services, credentials, or changing external
   state without a faithful local substitute.
3. Broad hidden or effectively unbounded behavior plus exact preservation of
   unrelated details, with explicitly end-to-end accept/reject feedback.
4. One unchanged artifact must satisfy independent grammars, parsers, runtimes,
   or environments, while checks expose only a joint accept/reject objective.
5. Bulk media or visual interpretation with no described numerical comparator
   and correctness resting on visual judgment.

Difficulty, visualization, specialization, or virtualization alone is not enough.
A small fixed visual artifact, scripted headless setup with direct checks, or
local scorer that distinguishes candidates should continue to later gates.
File-existence or format checks do not make long visual/media search a strong
oracle when the decisive event must still be found by inspecting many frames.

Gate 2: `research'

Choose `research' only when the primary deliverable is factual reconstruction
requiring multiple accessible sources, version/date alignment, provenance, or a
historical snapshot. A multi-record date/version ranking qualifies; one lookup or
research used only to build an artifact does not. If decisive sources are
explicitly inaccessible, choose `compact'.

Gate 3: `deep'

Choose `deep' only when a strong local oracle is explicitly described or can be
constructed from an exact deterministic reference, transformation, simulator, or
runtime supplied by the task, and at least one shape applies:

1. Latent structure or parameter recovery plus exact compatibility or a
   quantitative objective.
2. Correctness spans several interacting components, stages, or states and can be
   checked through exact invariants, reference behavior, or round trips.
3. Damaged structured-state recovery requiring competing format, ordering, or
   transformation hypotheses.
4. Constrained synthesis, compression, size reduction, or candidate search with
   coupled hard constraints and an executable scorer.
5. Nontrivial recovery, inference, or attack derived from many observations with
   exact candidate validation.
6. An inverse problem with a deterministic simulator or quantitative objective.
7. Candidate construction against a deterministic transformation or runtime,
   with multiple materially different mechanisms that can be executed and observed.
8. A severely constrained reimplementation of deterministic reference behavior,
   where exact candidate outputs can be compared locally despite language, size,
   compatibility, or dependency restrictions.
9. One source artifact must implement behavior while remaining valid under two
   or more independent grammars, compilers, or runtimes, each of which can be
   exercised separately for localized feedback.

Do not choose `deep' for terminology, long specifications, ordinary security or
reverse engineering, or hidden tests alone. Ordinary performance tuning and
ordinary state propagation through services, hooks, or deployments are
`standard'. A binary validator alone is insufficient, but a deterministic local
transformation plus executable behavior can support meaningful candidate
iteration even when the final verdict is binary. Prefer `standard' when each
candidate evaluation is described as expensive enough to consume a material
part of the watchdog and the work is mainly bounded tuning against one supplied
evaluator. Modifying an existing artifact against one supplied evaluator remains
`standard' merely because it combines a quantitative objective with preservation
constraints; `deep' additionally requires latent recovery, combinatorial search,
or several materially different candidate mechanisms. `deep' requires room for
several evidence-changing iterations.

Gate 4: `lean'

Choose `lean' only when all are explicit:

1. Bounded, mostly deterministic work.
2. At most two meaningful uncertainties.
3. A credible plan before broad exploration.
4. A direct local check covering nearly all decisive behavior.
5. About eight or fewer evidence-producing calls for inspection, implementation,
   and final validation.

Typical cases are a localized transformation; enumerable mechanical cleanup with
a deterministic final scan; a simple build or single-service setup with one main
uncertainty and a direct test; or bounded numerical work with direct value, shape,
or residual validation.

Do not choose `lean' when several coupled uncertainties require iterative
candidate comparison, or when recovering latent state requires materially
different hypotheses. Apply the `deep' gate first when a constructible exact or
quantitative reference makes those iterations useful. Also do not choose `lean'
for broad cleanup where unaffected content must be preserved exactly and the
set of true matches is not explicitly bounded; false-positive scope is a
separate uncertainty even when a final scan is deterministic.

Use `standard' when coordinating several schemas, formats, subsystems,
integration boundaries, correctness/performance constraints, or hidden edge-case
families. Also use it for multi-service state propagation, broad security
properties, builds combining source provenance with feature exclusion and
multiple runtime modes, corpus-wide cleanup with exact preservation of unaffected
content, or unstructured recovery with several evidence paths but only final
candidate checks.

Gate 5: `standard'

Choose `standard' for everything else, including unknown oracle quality. It is
the default for ordinary implementation, debugging, data processing, service
setup, bounded numerical work, and analysis with several actionable
uncertainties, and should normally be the plurality.
\end{artifacttextbox}

Listing~\ref{lst:tb2-hook} reproduces the budget-denial branch of
\texttt{\_pre\_tool\_use} and the complete functions that produce stage reminders
and compact text outputs. Dispatch, storage, audit logging, and other helper
routines are omitted from the listing but retained in the accompanying full files.

\begin{artifactbox}[lst:tb2-hook]{Medium: \texttt{runtime/hook.py} (excerpts)}
# Excerpt from _pre_tool_use: counter, budget, and denial branch
    tool_name = str(payload.get("tool_name") or "")
    state = _increment_tool_count(control_state)
    count = int(state["tool_count"])
    finalize = int(selection.get("finalize_after_tools") or 0)
    if finalize and count > finalize:
        _audit(
            "pre_tool_use",
            cwd=str(cwd),
            tool_name=tool_name,
            tool_count=count,
            decision="deny",
        )
        print(
            json.dumps(
                {
                    "hookSpecificOutput": {
                        "hookEventName": "PreToolUse",
                        "permissionDecision": "deny",
                        "permissionDecisionReason": (
                            "The task tool budget is exhausted. Preserve the current files, "
                            "stop calling tools, and give the best final answer from existing evidence."
                        ),
                    }
                }
            )
        )
        return 0

def _stage_context(count: int, selection: dict[str, Any]) -> str:
    economy = int(selection.get("economy_after_tools") or 0)
    strict = int(selection.get("strict_after_tools") or 0)
    reminder = int(selection.get("strict_reminder_every") or 8)
    if count == economy:
        return (
            "Economy phase starts now. Batch independent ls/rg/read/test checks, suppress "
            "process narration, and do not repeat unchanged checks. Do not trade away the "
            "behavioral validation or exact final-state checks required by the task."
        )
    if count >= strict and (count == strict or (count - strict) % reminder == 0):
        return (
            "Convergence phase: stop broad exploration. Only inspect evidence directly tied to the "
            "current failure or deliverable. If a concrete failure, unexplained instability, or "
            "mutated deliverable remains, repair it rather than finalizing because the phase changed. "
            "Avoid irreversible cleanup or broad state changes not required by the task, and keep "
            "temporary verification artifacts separate from final deliverables. "
            "Then recheck exact paths, interfaces, behavior, and final state once and finish."
        )
    return ""

def _compact_text(text: str, *, head_chars: int, tail_chars: int) -> str:
    lines = text.splitlines()
    deduped: list[str] = []
    index = 0
    while index < len(lines):
        next_index = index + 1
        while next_index < len(lines) and lines[next_index] == lines[index]:
            next_index += 1
        repeats = next_index - index
        deduped.append(lines[index])
        if repeats > 2:
            deduped.append(f"[previous line repeated {repeats - 1} more times]")
        elif repeats == 2:
            deduped.append(lines[index])
        index = next_index
    compact = "\n".join(deduped)
    if len(compact) <= head_chars + tail_chars:
        return compact
    omitted = len(compact) - head_chars - tail_chars
    return (
        compact[:head_chars]
        + f"\n... [{omitted} characters omitted; use the raw path for a targeted range] ...\n"
        + compact[-tail_chars:]
    )
\end{artifactbox}

The policy below specifies the execution profiles, routing rules, and
output-retention settings.
\texttt{history\_token\_limit} is a
per-output history limit, not the model context-window size.
The rubric guides profile selection. The runtime hook enforces the tool-call
limits specified by the policy's \texttt{finalize\_after\_tools} values.
Tool thresholds count attempted calls.

\begin{artifactbox}[lst:tb2-reference-policy]{Medium: \texttt{policy.json} (selected fields)}
{
  "default_profile": "standard",
  "tool_output": {
    "history_token_limit": 900,
    "max_chars": 4000,
    "head_chars": 2400,
    "tail_chars": 900,
    "rtk_enabled": true
  },
  "profiles": {
    "compact": {
      "reasoning_effort": "low",
      "economy_after_tools": 2,
      "strict_after_tools": 4,
      "finalize_after_tools": 14,
      "strict_reminder_every": 2
    },
    "lean": {
      "reasoning_effort": "low",
      "economy_after_tools": 6,
      "strict_after_tools": 10,
      "finalize_after_tools": 22,
      "strict_reminder_every": 4
    },
    "standard": {
      "reasoning_effort": "medium",
      "economy_after_tools": 14,
      "strict_after_tools": 26,
      "finalize_after_tools": 44,
      "strict_reminder_every": 8
    },
    "deep": {
      "reasoning_effort": "high",
      "economy_after_tools": 16,
      "strict_after_tools": 30,
      "finalize_after_tools": 54,
      "strict_reminder_every": 8
    },
    "research": {
      "reasoning_effort": "high",
      "economy_after_tools": 8,
      "strict_after_tools": 20,
      "finalize_after_tools": 46,
      "strict_reminder_every": 6
    }
  },
  "classifier": {
    "type": "model",
    "model": "gpt-5.5",
    "reasoning_effort": "medium",
    "transport": "responses",
    "fallback_profile": "standard",
    "rubric_file": "profile_rubric.md",
    "timeout_s": 120,
    "max_attempts": 3,
    "max_output_tokens": 2048,
    "samples": 1,
    "max_task_chars": 24000
  },
  "confidence_overrides": [],
  "rules": []
}
\end{artifactbox}


\subsubsection{Lite Mode}
\label{app:tb2-lite}
The lite mode retains the shared rubric but adjusts the tool budgets and
reasoning settings through \texttt{policy.json}.
The \texttt{deep} and \texttt{research} profiles use medium reasoning effort.
The executable settings below therefore take precedence over the effort
descriptions inherited in the rubric. Listing~\ref{lst:tb2-lite-policy} gives
the complete policy diff from the medium configuration; all other fields are
unchanged.

\begin{artifactbox}[lst:tb2-lite-policy]{Lite: \texttt{runtime/policy.json} (diff vs. medium)}
--- medium/runtime/policy.json
+++ lite/runtime/policy.json
@@ -40,5 +40,5 @@
       "economy_after_tools": 2,
       "strict_after_tools": 4,
-      "finalize_after_tools": 14,
+      "finalize_after_tools": 8,
       "strict_reminder_every": 2
     },
@@ -47,5 +47,5 @@
       "economy_after_tools": 6,
       "strict_after_tools": 10,
-      "finalize_after_tools": 22,
+      "finalize_after_tools": 16,
       "strict_reminder_every": 4
     },
@@ -54,19 +54,19 @@
       "economy_after_tools": 14,
       "strict_after_tools": 26,
-      "finalize_after_tools": 44,
+      "finalize_after_tools": 38,
       "strict_reminder_every": 8
     },
     "deep": {
-      "reasoning_effort": "high",
+      "reasoning_effort": "medium",
       "economy_after_tools": 16,
       "strict_after_tools": 30,
-      "finalize_after_tools": 54,
+      "finalize_after_tools": 48,
       "strict_reminder_every": 8
     },
     "research": {
-      "reasoning_effort": "high",
+      "reasoning_effort": "medium",
       "economy_after_tools": 8,
       "strict_after_tools": 20,
-      "finalize_after_tools": 46,
+      "finalize_after_tools": 40,
       "strict_reminder_every": 6
     }
\end{artifactbox}

\subsubsection{Heavy Mode}
\label{app:tb2-heavy}
The heavy mode uses GPT-5.5 with low, medium, or high reasoning effort,
depending on the selected profile, and allows larger tool budgets than medium.
It has the same instruction and hook files as medium.
Only the routing rubric and the policy's \texttt{profiles} and
\texttt{session\_context} fields differ. The rubric diff below is
relative to Listing~\ref{lst:tb2-rubric}; the changed policy fields follow.

\begin{artifactbox}[lst:tb2-heavy-rubric]{Heavy: \texttt{profile\_rubric.md} (diff vs. medium)}
--- medium/runtime/profile_rubric.md
+++ heavy/runtime/profile_rubric.md
@@ -7,5 +7,5 @@
-- `compact': low reasoning; converge near 4 tool calls and stop near 8.
-- `lean': low reasoning; converge near 10 calls and stop near 16.
-- `standard': medium reasoning; converge near 26 calls and stop near 38.
-- `deep': high reasoning; converge near 30 calls and stop near 48.
-- `research': high reasoning with about 40 calls for source reconstruction.
+- `compact': low reasoning; converge near 8 tool calls and stop near 14.
+- `lean': medium reasoning; converge near 22 calls and stop near 36.
+- `standard': high reasoning; converge near 38 calls and stop near 60.
+- `deep': high reasoning; converge near 42 calls and stop near 68.
+- `research': high reasoning with up to about 60 calls for source reconstruction.
@@ -113 +113 @@
-5. About eight or fewer evidence-producing calls for inspection, implementation,
+5. About twelve or fewer evidence-producing calls for inspection, implementation,
\end{artifactbox}

\begin{artifacttextbox}[lst:tb2-heavy-session]{Heavy: \texttt{policy.json}: \texttt{session\_context}}
Harness guidance is already loaded; do not inventory harness/. Extract exact artifacts, paths, behavior, integrity restrictions, public interfaces, and final state from task/prompt.md. Use its host-to-container pattern. For black-box work use only public interfaces and implement general behavior. When the task names an API, wrapper, package, revision, runtime, or service, exercise that named interface first and do not silently replace it with a lower-level dependency, mock, proxy, or look-alike behavior; inspect the public signature or implementation after an interface error. A source-only or single-artifact deliverable should not acquire undeclared runtime helpers. Start with the smallest discriminating probe and batch independent checks. Bound downloads, external interaction, and potentially blocking commands well below the task watchdog; reserve time for final repair and verification. Validate requested behavior, not only superficial readiness signals. For multi-stage services and persistent state, test from a fresh external entry point and check repeated-operation behavior, defaults, permissions, and final ownership instead of trusting internal readiness. For numerical, transformation, or recovery work, validate against an independent reference, invariant, round trip, or residual rather than only reusing the implementation's own assumptions. Run mutating workflow checks on disposable fixtures when possible; if the user is meant to perform the same workflow later, do not leave test data or state that collides with its documented preconditions. Preserve the real requested implementation and durable service state while removing only temporary verification artifacts. Keep temporary verification artifacts separate from final deliverables. Before finishing, reread the public task once, inspect the actual final artifact or durable state, and compare every explicit requirement against independent evidence. Avoid narration and unchanged rereads; do not skip necessary validation.
\end{artifacttextbox}

\begin{artifactbox}[lst:tb2-heavy-policy]{Heavy: \texttt{policy.json}: \texttt{profiles}}
{
  "profiles": {
    "compact": {
      "reasoning_effort": "low",
      "economy_after_tools": 4,
      "strict_after_tools": 8,
      "finalize_after_tools": 14,
      "strict_reminder_every": 3
    },
    "lean": {
      "reasoning_effort": "medium",
      "economy_after_tools": 10,
      "strict_after_tools": 22,
      "finalize_after_tools": 36,
      "strict_reminder_every": 4
    },
    "standard": {
      "reasoning_effort": "high",
      "economy_after_tools": 18,
      "strict_after_tools": 38,
      "finalize_after_tools": 60,
      "strict_reminder_every": 8
    },
    "deep": {
      "reasoning_effort": "high",
      "economy_after_tools": 20,
      "strict_after_tools": 42,
      "finalize_after_tools": 68,
      "strict_reminder_every": 8
    },
    "research": {
      "reasoning_effort": "high",
      "economy_after_tools": 16,
      "strict_after_tools": 36,
      "finalize_after_tools": 60,
      "strict_reminder_every": 8
    }
  }
}
\end{artifactbox}

\subsection{SWE-Bench Pro}
\label{app:swebench-pro-artifacts}

The medium mode harness uses GPT-5.5 and Codex.
A separate GPT-5.5 call uses the rubric to select one of five reasoning and tool-budget
profiles from the public task description. The repository-repair workflow first
extracts the contract, then
checks the implementation and available references, makes a focused patch, and
verifies task-shaped behavior.
Listing~\ref{lst:swe-medium-layout} shows the stored harness files. The evaluator
configures Codex to load the instructions and register the hook. The
\texttt{repair-verify} script provides basic code checks, while a checklist
guides task-specific verification.

\begin{artifactbox}[lst:swe-medium-layout]{SWE-Bench Pro medium-mode harness files}
README.md
runtime/
    model_instructions.md
    profile_rubric.md
    policy.json
    hook.py
bin/
    repair-verify
checklists/
    contract-verification.md
\end{artifactbox}

\begin{artifacttextbox}[lst:swe-medium-instructions]{SWE-Bench Pro medium: \texttt{model\_instructions.md}}
You are Codex, a non-interactive software agent. Work autonomously until the
user's task is complete. Follow the task's integrity, safety, path, and output
requirements exactly. Use the available tools to inspect, edit, execute, and
verify; batch independent checks and avoid repeating unchanged reads. Preserve
unrelated files and avoid destructive operations outside the requested scope.
Do not replace a requested implementation or service with a mock, proxy, or
look-alike unless the task explicitly permits that substitution.
Do not spawn other agents or ask the user to perform routine steps. Keep
progress messages brief. Before finishing, leave every requested artifact at
its exact path and run the strongest allowed task-shaped verification. Report
the outcome concisely, including any genuine remaining uncertainty.
\end{artifacttextbox}

\begin{artifacttextbox}[lst:swe-medium-session]{SWE-Bench Pro medium: \texttt{policy.json}: \texttt{session\_context}}
The repair workflow is already loaded; do not spend tool calls inventorying harness/. First establish the repository root under task/ and keep later commands rooted there. Read task/prompt.md as a contract and keep a concise ledger of exact public names, files, payloads, defaults, error behavior, compatibility requirements, and stated edge cases. Inspect the local implementation and nearby tests first. Look for an existing reference when the task sounds like a known fix: local Git history, tags, upstream commits, tests, fixtures, or nearby implementation patterns. Prefer adapting proven behavior over guessing. Trace the real execution and data path before editing, then make the smallest production-source change that satisfies the contract. Start with narrow discriminating probes, batch independent reads and searches, and avoid unchanged rereads. Reserve enough budget for a post-change task-shaped test, investigation of any concrete failure, and one final diff/contract review; compilation alone is insufficient. Use harness/bin/repair-verify only when its general checks are useful. Keep generated files, caches, and temporary tests out of the final patch. Before finishing, inspect the diff and re-check every ledger item. Preserve unrelated behavior and do not replace requested behavior with a mock, proxy, or look-alike implementation.
\end{artifacttextbox}

\newpage
Relative to the Terminal-Bench~2 medium rubric in Listing~\ref{lst:tb2-rubric},
the SWE-Bench Pro rubric changes the profile budgets and the \texttt{lean}
selection gate. The diff below shows these changes.

\begin{artifactbox}[lst:swe-medium-rubric]{SWE-Bench Pro medium: \texttt{profile\_rubric.md} (diff vs. Terminal-Bench~2)}
--- tb2/medium/runtime/profile_rubric.md
+++ swebench_pro/medium/runtime/profile_rubric.md
@@ -7,6 +7,6 @@
 - `compact': low reasoning; converge near 4 tool calls and stop near 8.
-- `lean': low reasoning; converge near 10 calls and stop near 16.
-- `standard': medium reasoning; converge near 26 calls and stop near 38.
-- `deep': high reasoning; converge near 30 calls and stop near 48.
-- `research': high reasoning with about 40 calls for source reconstruction.
+- `lean': low reasoning; converge near 16 calls and stop near 24.
+- `standard': medium reasoning; converge near 42 calls and stop near 64.
+- `deep': high reasoning; converge near 48 calls and stop near 72.
+- `research': high reasoning with up to about 64 calls for source reconstruction.

@@ -112,4 +112,4 @@
 4. A direct local check covering nearly all decisive behavior.
-5. About eight or fewer evidence-producing calls for inspection, implementation,
-   and final validation.
+5. About twelve or fewer evidence-producing calls for inspection,
+   implementation, failure diagnosis, and final validation.

@@ -119,2 +119,8 @@
 or residual validation.
+
+For an existing-code repair, use `lean' only when the described behavior is
+confined to one narrow implementation surface and one focused local check can
+cover both the requested success path and its stated edge cases. A short prompt
+or a small-looking diff is not enough when compatibility, lifecycle, persistence,
+concurrency, or cross-layer behavior remains coupled.
\end{artifactbox}

The selected policy fields below specify routing, profile effort and tool
stages, confidence handling, and output retention. Tool thresholds count attempted calls, while
\texttt{history\_token\_limit} limits each retained tool output, not the model
context window.

\begin{artifactbox}[lst:swe-medium-policy]{SWE-Bench Pro medium: \texttt{policy.json} (selected fields)}
{
  "default_profile": "standard",
  "tool_output": {
    "history_token_limit": 1400,
    "max_chars": 6000,
    "head_chars": 3600,
    "tail_chars": 1400,
    "rtk_enabled": true
  },
  "profiles": {
    "compact": {
      "reasoning_effort": "low",
      "economy_after_tools": 2,
      "strict_after_tools": 4,
      "finalize_after_tools": 8,
      "strict_reminder_every": 2
    },
    "lean": {
      "reasoning_effort": "low",
      "economy_after_tools": 8,
      "strict_after_tools": 16,
      "finalize_after_tools": 24,
      "strict_reminder_every": 4
    },
    "standard": {
      "reasoning_effort": "medium",
      "economy_after_tools": 20,
      "strict_after_tools": 42,
      "finalize_after_tools": 64,
      "strict_reminder_every": 8
    },
    "deep": {
      "reasoning_effort": "high",
      "economy_after_tools": 24,
      "strict_after_tools": 48,
      "finalize_after_tools": 72,
      "strict_reminder_every": 8
    },
    "research": {
      "reasoning_effort": "high",
      "economy_after_tools": 16,
      "strict_after_tools": 40,
      "finalize_after_tools": 64,
      "strict_reminder_every": 8
    }
  },
  "classifier": {
    "type": "model",
    "model": "gpt-5.5",
    "reasoning_effort": "medium",
    "fallback_profile": "standard",
    "rubric_file": "profile_rubric.md",
    "timeout_s": 120,
    "max_attempts": 3,
    "max_task_chars": 24000
  },
  "confidence_overrides": [
    {
      "name": "uncertain_lean_to_standard",
      "source_profile": "lean",
      "below": 0.9,
      "reasoning_effort": "medium",
      "economy_after_tools": 20,
      "strict_after_tools": 42,
      "finalize_after_tools": 64,
      "strict_reminder_every": 8
    }
  ],
  "rules": []
}
\end{artifactbox}

The hook excerpts reproduce the complete budget gate, stage-reminder,
and text-compaction functions. Dispatch, audit, storage, and the remaining helper
routines are retained in the accompanying complete \texttt{hook.py}.

\begin{artifactbox}[lst:swe-medium-hook]{SWE-Bench Pro medium: \texttt{hook.py} (excerpts)}
def _pre_tool_use(
    payload: dict[str, Any],
    cwd: Path,
    control_state: Path,
    selection: dict[str, Any],
) -> int:
    tool_name = str(payload.get("tool_name") or "")
    state = _increment_tool_count(control_state)
    count = int(state["tool_count"])
    finalize = int(selection.get("finalize_after_tools") or 0)
    if finalize and count > finalize:
        _audit(
            "pre_tool_use",
            cwd=str(cwd),
            tool_name=tool_name,
            tool_count=count,
            decision="deny",
        )
        print(
            json.dumps(
                {
                    "hookSpecificOutput": {
                        "hookEventName": "PreToolUse",
                        "permissionDecision": "deny",
                        "permissionDecisionReason": (
                            "The task tool budget is exhausted. Preserve the current files, "
                            "stop calling tools, and give the best final answer from existing evidence."
                        ),
                    }
                }
            )
        )
        return 0

    context = _stage_context(count, selection)
    updated_input = _rtk_rewrite(payload, selection) if tool_name == "Bash" else None
    _audit(
        "pre_tool_use",
        cwd=str(cwd),
        tool_name=tool_name,
        tool_count=count,
        stage=_stage_name(count, selection),
        input_rewritten=updated_input is not None,
    )
    if context or updated_input is not None:
        specific: dict[str, Any] = {"hookEventName": "PreToolUse"}
        if context:
            specific["additionalContext"] = context
        if updated_input is not None:
            specific["permissionDecision"] = "allow"
            specific["updatedInput"] = updated_input
        print(json.dumps({"hookSpecificOutput": specific}))
    return 0

def _stage_context(count: int, selection: dict[str, Any]) -> str:
    economy = int(selection.get("economy_after_tools") or 0)
    strict = int(selection.get("strict_after_tools") or 0)
    reminder = int(selection.get("strict_reminder_every") or 8)
    if count == economy:
        return (
            "Economy phase starts now. Batch independent ls/rg/read/test checks, suppress "
            "process narration, and do not repeat unchanged checks. Do not trade away the "
            "behavioral validation or exact final-state checks required by the task."
        )
    if count >= strict and (count == strict or (count - strict) % reminder == 0):
        return (
            "Convergence phase: stop broad exploration. Only inspect evidence directly tied to the "
            "current failure or deliverable. If a concrete failure, unexplained instability, or "
            "mutated deliverable remains, repair it rather than finalizing because the phase changed. "
            "Avoid irreversible cleanup or broad state changes not required by the task, and keep "
            "temporary verification artifacts separate from final deliverables. "
            "Then recheck exact paths, interfaces, behavior, and final state once and finish."
        )
    return ""

def _compact_text(text: str, *, head_chars: int, tail_chars: int) -> str:
    lines = text.splitlines()
    deduped: list[str] = []
    index = 0
    while index < len(lines):
        next_index = index + 1
        while next_index < len(lines) and lines[next_index] == lines[index]:
            next_index += 1
        repeats = next_index - index
        deduped.append(lines[index])
        if repeats > 2:
            deduped.append(f"[previous line repeated {repeats - 1} more times]")
        elif repeats == 2:
            deduped.append(lines[index])
        index = next_index
    compact = "\n".join(deduped)
    if len(compact) <= head_chars + tail_chars:
        return compact
    omitted = len(compact) - head_chars - tail_chars
    return (
        compact[:head_chars]
        + f"\n... [{omitted} characters omitted; use the raw path for a targeted range] ...\n"
        + compact[-tail_chars:]
    )
\end{artifactbox}

\subsection{ProgramBench}
\label{app:programbench-artifacts}

The ProgramBench medium-mode harness uses a fixed GPT-5.5 high-effort profile
instead of task-adaptive routing. The workflow builds an
independent program from bundled documentation and permitted black-box probes,
validates it against the public reference, and reserves time for a clean offline
rebuild.
Listing~\ref{lst:program-medium-layout} shows the stored harness files. This
fixed-profile harness has no routing rubric. The evaluator configures Codex to
load the instruction file and execute the budget hook.

\begin{artifactbox}[lst:program-medium-layout]{ProgramBench medium-mode harness files}
README.md
runtime/
    model_instructions.md
    policy.json
    hook.py
\end{artifactbox}

\begin{artifacttextbox}[lst:program-medium-instructions]{ProgramBench medium: \texttt{model\_instructions.md}}
# ProgramBench Reconstruction Harness

Use this harness as a task-agnostic control loop for rebuilding a program from
documentation and black-box observations. It supplements the benchmark rules;
it never permits source retrieval, internet access, decompilation, tracing, or
reuse of the reference executable.

## Evidence-first workflow

1. Read all bundled documentation and turn it into a compact behavior ledger:
   commands, flags, inputs, outputs, exit status, file effects, defaults, error
   behavior, formatting, and explicitly documented edge cases.
2. Establish a runnable skeleton early. Create `compile.sh', build a minimal
   independent executable, and confirm the complete build/run cycle before
   investing in secondary behavior.
3. Probe the reference through its normal public interface. Use small,
   discriminating experiments that separate plausible behaviors; record the
   observed input, output, status, and state change instead of relying on
   assumptions based on the program name.
4. Prioritize breadth before polish. Implement the documented happy paths and
   common input classes, then error handling, boundary cases, formatting, and
   low-frequency options. Keep a list of remaining unsupported behavior.
5. Compare your implementation and the reference on the same locally generated
   cases. When they differ, reduce the case and fix the underlying general rule
   rather than adding a one-off special case.
6. Rebuild from a clean state periodically. A solution that works only because
   of cached artifacts, undeclared tools, or files outside the submission is
   not complete.
7. Before finalizing, verify that `compile.sh' is executable, produces
   `./executable' without network access, and that representative success and
   failure probes agree with the reference.

## Efficiency rules

- Batch independent documentation reads and related probes when practical.
- Prefer high-information probes over broad random exploration.
- Keep outputs bounded and summarize findings in the behavior ledger.
- Stop repeating probes once a stable rule has been established.
- Reserve the final portion of the budget for a clean rebuild and end-to-end
  validation rather than continued exploration.

## Reference lifecycle

Keep the black-box reference available until differential validation is finished.
Develop a separately built candidate at another path, with the same executable
basename where relevant. Do not overwrite the reference during implementation.
If a permitted move is necessary, retain the execute-only reference on the same
filesystem without reading/copying its bytes or changing its permissions.
The reference is only a development oracle: never include it in the submission
or invoke it from the submitted implementation. Self-authored expected outputs
do not replace an actual reference-versus-candidate comparison. Build the final
deliverable from the validated independent source after the last comparison.
\end{artifacttextbox}

The policy's \texttt{history\_token\_limit} is a per-output history-retention limit, not
the model context size. The hook applies staged tool budgets; its hard-stop
threshold is the finalization threshold plus the larger of eight and the
reminder interval.

\begin{artifactbox}[lst:program-medium-policy]{ProgramBench medium: \texttt{policy.json}}
{
  "version": 1,
  "default_profile": "heavy",
  "session_context": "Follow the reconstruction workflow: read bundled documentation, preserve the reference for permitted black-box comparisons, build an independent implementation, and reserve time for an offline rebuild and end-to-end validation.",
  "codex": {
    "model_instructions_file": "model_instructions.md",
    "model_verbosity": "low",
    "disabled_features": [
      "apps",
      "auth_elicitation",
      "browser_use",
      "browser_use_external",
      "browser_use_full_cdp_access",
      "computer_use",
      "fast_mode",
      "goals",
      "image_generation",
      "in_app_browser",
      "multi_agent",
      "personality",
      "plugins",
      "remote_plugin",
      "shell_snapshot",
      "skill_mcp_dependency_install",
      "tool_call_mcp_elicitation",
      "tool_suggest",
      "workspace_dependencies"
    ]
  },
  "tool_output": {
    "history_token_limit": 3000,
    "max_chars": 4500,
    "head_chars": 2700,
    "tail_chars": 1100,
    "rtk_enabled": false
  },
  "profiles": {
    "heavy": {
      "reasoning_effort": "high",
      "economy_after_tools": 36,
      "strict_after_tools": 60,
      "finalize_after_tools": 76,
      "strict_reminder_every": 8
    }
  },
  "rules": []
}
\end{artifactbox}

\begin{artifactbox}[lst:program-medium-hook]{ProgramBench medium: \texttt{hook.py}}
#!/usr/bin/env python3
"""Task-agnostic staged budget hook for ProgramBench Codex runs."""

from __future__ import annotations

import fcntl
import json
import os
from pathlib import Path
import sys


def _selection() -> dict:
    path = os.environ.get("RHO_RUNTIME_SELECTION_PATH", "").strip()
    if not path:
        return {}
    try:
        value = json.loads(Path(path).read_text(encoding="utf-8"))
    except (OSError, json.JSONDecodeError):
        return {}
    return value if isinstance(value, dict) else {}


def _next_count(cwd: Path) -> int:
    configured = os.environ.get("RHO_RUNTIME_STATE_ROOT", "").strip()
    state_dir = Path(configured).resolve() if configured else cwd / ".rho_runtime"
    state_dir.mkdir(parents=True, exist_ok=True)
    path = state_dir / "tool-count.json"
    lock_path = state_dir / "tool-count.lock"
    with lock_path.open("a+") as lock:
        fcntl.flock(lock, fcntl.LOCK_EX)
        try:
            value = json.loads(path.read_text(encoding="utf-8"))
            count = int(value.get("tool_count") or 0) + 1
        except (OSError, ValueError, TypeError, json.JSONDecodeError):
            count = 1
        temp = path.with_suffix(".tmp")
        temp.write_text(json.dumps({"tool_count": count}) + "\n", encoding="utf-8")
        temp.replace(path)
        return count


def _context(event: str, text: str) -> None:
    print(json.dumps({"hookSpecificOutput": {"hookEventName": event, "additionalContext": text}}))


def main() -> int:
    payload = json.load(sys.stdin)
    event = str(payload.get("hook_event_name") or "")
    selection = _selection()
    if event == "SessionStart":
        context = str(selection.get("session_context") or "").strip()
        if context:
            _context("SessionStart", context)
        return 0
    if event != "PreToolUse":
        return 0
    cwd = Path(str(payload.get("cwd") or os.getcwd())).resolve()
    count = _next_count(cwd)
    economy = int(selection.get("economy_after_tools") or 0)
    strict = int(selection.get("strict_after_tools") or 0)
    finalize = int(selection.get("finalize_after_tools") or 0)
    reminder = int(selection.get("strict_reminder_every") or 1)
    hard_stop = finalize + max(8, reminder) if finalize else 0
    if hard_stop and count > hard_stop:
        print(
            json.dumps(
                {
                    "hookSpecificOutput": {
                        "hookEventName": "PreToolUse",
                        "permissionDecision": "deny",
                        "permissionDecisionReason": (
                            "The exploration budget is exhausted. Preserve the current files, "
                            "finish from existing evidence, and do not call more tools."
                        ),
                    }
                }
            )
        )
        return 0
    if finalize and count >= finalize and (
        count == finalize or (count - finalize) % max(4, reminder) == 0
    ):
        _context(
            "PreToolUse",
            "Finalization stage: stop adding semantic scope. Use the remaining small grace "
            "window only for implementation completion, a clean build, a focused check, "
            "artifact cleanup, or submission. Then finish.",
        )
    elif strict and count >= strict and (count == strict or (count - strict) % reminder == 0):
        _context(
            "PreToolUse",
            "Convergence stage: stop broad exploration. Revisit the public behavior ledger, "
            "use only probes that distinguish a live hypothesis or repair a concrete mismatch, "
            "and preserve enough budget for an offline rebuild and final contract check.",
        )
    elif economy and count == economy:
        _context(
            "PreToolUse",
            "Economy stage: consolidate the behavior ledger, batch related checks, and cover "
            "remaining documented semantic families before low-frequency polish.",
        )
    return 0


if __name__ == "__main__":
    raise SystemExit(main())
\end{artifactbox}

\end{document}